\documentclass[10.5pt,letter]{elsarticle}  
\usepackage[letterpaper, total={6in, 8in}]{geometry}

\usepackage{amsmath,amssymb,amsmath,bm}
\usepackage{color}
\usepackage{times}
\usepackage{url,flushend,multirow,booktabs}
\usepackage[ruled,linesnumbered]{algorithm2e}  
\usepackage{graphicx}
\usepackage{subfigure}
\usepackage{float}
\usepackage{hyphenat}
\usepackage[colorlinks,citecolor=green,urlcolor=blue,bookmarks=false,hypertexnames=true]{hyperref}

\journal{arXiv}

\usepackage{numcompress}
\newcommand{\eg}{e.g.}
\newcommand{\ie}{i.e.}

\newcommand{\etal}{\textit{et al.}}
\newcommand{\aka}{\textit{a.k.a.}}

\DeclareFixedFont{\mf}{OT1}{ptm}{m}{n}{10pt}
\DeclareFixedFont{\mfb}{OT1}{ptm}{bx}{n}{10pt}

\begin{document}

\begin{frontmatter}

\title{Fully Transformer-Equipped Architecture for End-to-End Referring Video Object Segmentation}


\author[add1,add2]{Ping~Li\corref{mark1}}
\cortext[mark1]{Corresponding author}
\ead{patriclouis.lee@gmail.com}
\author[add1]{Yu~Zhang}  
\author[add3]{Li~Yuan} 
\author[add1]{Xianghua~Xu}  

\address[add1]{School of Computer Science and Technology, Hangzhou Dianzi University, Hangzhou, China}
\address[add2]{Guangdong Laboratory of Artificial Intelligence and Digital Economy (SZ), Shenzhen, China}
\address[add3]{School of Electronic and Computer Engineering, Peking University, Beijing, China}

\begin{abstract}
    Referring Video Object Segmentation (RVOS) requires segmenting the object in video referred by a natural language query. Existing methods mainly rely on sophisticated pipelines to tackle such cross-modal task, and do not explicitly model the object-level spatial context which plays an important role in locating the referred object. Therefore, we propose an end-to-end RVOS framework completely built upon transformers, termed \textit{Fully Transformer-Equipped Architecture} (FTEA), which treats the RVOS task as a mask sequence learning problem and regards all the objects in video as candidate objects. Given a video clip with a text query, the visual-textual features are yielded by encoder, while the corresponding pixel-level and word-level features are aligned in terms of semantic similarity. To capture the object-level spatial context, we have developed the Stacked Transformer, which individually characterizes the visual appearance of each candidate object, whose feature map is decoded to the binary mask sequence in order directly. Finally, the model finds the best matching between mask sequence and text query. In addition, to diversify the generated masks for candidate objects, we impose a diversity loss on the model for capturing more accurate mask of the referred object. Empirical studies have shown the superiority of the proposed method on three benchmarks, e.g., FETA achieves 45.1\% and 38.7\% in terms of mAP on A2D Sentences (3782 videos) and J-HMDB Sentences (928 videos), respectively; it achieves 56.6\% in terms of $\mathcal{J\&F}$ on Ref-YouTube-VOS (3975 videos and 7451 objects). Particularly, compared to the best candidate method, it has a gain of 2.1\% and 3.2\% in terms of P$@$0.5 on the former two, respectively, while it has a gain of 2.9\% in terms of $\mathcal{J}$ on the latter one.
\end{abstract}   


\end{frontmatter}


\section{Introduction}
\label{sec1:intro}
Referring Video Object Segmentation (RVOS) bridges the semantic gap between text description and video content, by yielding the pixel-level object mask sequence, which matches with the referred target in a sentence (\aka, text query or referring expression). Generally, RVOS has widespread application fields, such as human-robot interaction \cite{qi-cvpr2020-robotics} and language-based video editing \cite{fu-arxiv2021-video-edit}. Unlike popular semi-supervised video object segmentation \cite{hu-cvpr2021-lcm,cheng-nips2021-stcn} that provides the first-frame mask, there is none of pixel labeling and the visual-linguistic cross-modality understanding is required for RVOS, which makes it much more challenging.

\begin{figure*}[!tb]
	\centering
	\includegraphics[width=0.9\textwidth,height=0.32\textwidth]{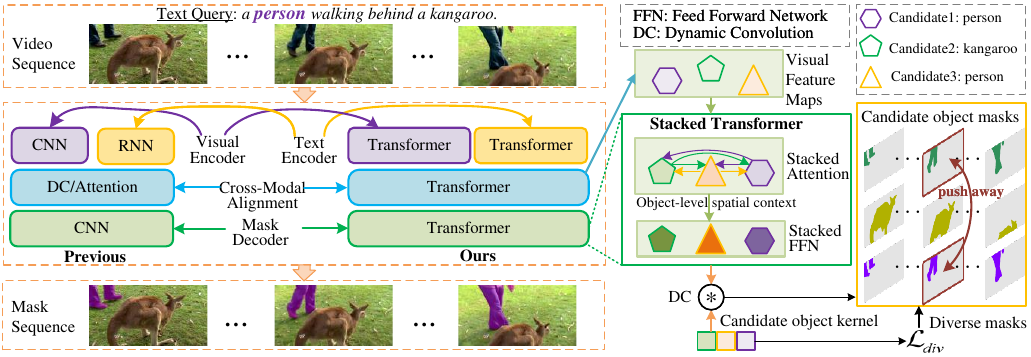}
	\caption{Overall pipeline of our model. It adopts an end-to-end framework completely built upon transformers, and the Stacked Transformer with diversity loss is developed for decoding mask sequence of the referred object (person).}
	\label{fig:motivation}
	\vspace{-0.5pt}
\end{figure*}

\textbf{Research Objectives}. We have two primary goals: 1) matching the objects with the text query by cross-modal feature alignment; 2) locating the pixel regions of candidate objects by learning promising feature representation which captures the spatiotemporal pixel relations among frames.

Essentially, RVOS requires understanding the scene and the objects in video while matching objects with text query. The challenge is there are usually more than one object, and it desires modeling the relative positions of objects and action interactions. However, most existing methods regard RVOS as a pixel-wise classification problem, classifying each pixel of input video into two categories, namely target or background. Hence, the non-referred objects may be treated as background, adding difficulty to modeling the object spatial relations. Here, we argue that RVOS is an object-wise rather than pixel-wise classification problem. For example, there are a person and a kangaroo in Figure~\ref{fig:motivation}, and the referring expression is ``\textit{a person walking behind the kangaroo}''; thus the model needs to capture both the person and the kangaroo at the same time while modeling the relations of the two objects, and then segment the target object (person), \ie, the subject of the referring sentence. Existing pixel-wise methods \cite{gavrilyuk-cvpr2018-aavs,seo-eccv2020-urvos,wang-aaai2020-cmdn} fail to model the spatial relations at object level, thus degrading the performance.

To overcome the above limitation, we rethink the RVOS as an object-wise classification problem, and adopt the paradigm of mask classification \cite{cheng-nips2021-maskformer} to construct a novel object-wise framework. The basic idea is to generate binary masks of multiple candidate objects, and then select the best candidate object mask as the final prediction. As shown in Figure~\ref{fig:motivation}, the video and the referring expression involve two candidate objects, ``person'' and ``kangaroo''.  Our method aims at generating masks for all the objects and finally selecting the binary mask corresponding to ``person'' as the target mask. In the mask classification paradigm, the goal is to capture all candidate objects in video, while effectively understanding the potential relations among candidate objects. 

To locate the pixel regions of candidate objects, a natural way is to obtain their features individually. However, existing methods \cite{botach-cvpr2022-mttr,seo-eccv2020-urvos} adopt convolution and upsampling operations to decode frame features into masks. But the feature representation of pixel region mainly uses the convolution operations to locally capture the spatial context to obtain image-level feature, which lacks global relation modeling and cannot independently characterize each candidate object. Thus, we propose a transformer-based object decoding network to globally capture the spatial context and obtain independent candidate object feature maps. 

To this end, we propose a Fully Transformer-Equipped Architecture (FTEA) for RVOS task, and the overall pipeline is illustrated in Figure~\ref{fig:motivation}. Given a video and a text query (left top), FTEA outputs mask sequence matching with the referred target (person in purple at the left bottom). FTEA is composed of Visual Encoder, Text Encoder, Cross-Modal Alignment module, and Mask Decoder. Unlike previous complex pipelines \cite{liu-pami2021-cmpc, ye-pami2021-cmsa, wang-aaai2020-cmdn, seo-eccv2020-urvos} unifying both CNNs (Convolution Neural Networks) \cite{carreira-cvpr2017-i3d, szegedy-cvpr2016-inceptionv3} and RNNs (Recurrent Neural Networks) \cite{chung-arxiv2014-gru, hochreiter-nc-1997-lstm}), our framework is completely dependent upon efficient transformers including visual and text feature encoding. More importantly, we propose Stacked Transformer based Mask Decoder to decode frame feature maps to mask sequences by designing two modules, \ie, Stacked Attention (SA) and Stacked Feed Forward Network (SFFN). It employs a progressive object learning strategy for capturing the object-level spatial context, and utilize dynamic convolution with the imposed diversity loss on candidate object kernels, in order to make candidate object masks diverse. In Mask Decoder, one linear layer follows Stacked Transformer to join the dynamic convolution with candidate object kernels to produce corresponding masks. The referred object matches with the mask sequence with the highest sum of referring scores, and these scores are from Cross-Modal Alignment module.

The former SA module takes low-level feature maps and cross-modal alignment features as input, trying to capture the appearance cues, such as edges and contours, and thus locate the region of interest relating with the referred object. To obtain independent candidate object feature maps, SA applies candidate object kernels on pixel-wise features of the image, and object kernels encode the object-level property such as location or action. The latter SFFN module, placed behind SA, divides feature map channels into groups, whose number equals the number of candidate objects, resulting in group sparsity of object relations during group linear mapping. Each group of feature maps is learned to reveal the latent pattern of one candidate object. Due to the sparsely group-wise strategy, computational cost is largely reduced compared to densely channel-wise way adopted by common convolution. 

In mask classification paradigm, the number of candidate object masks is often much greater than that of objects in video, and there may be a phenomenon that multiple candidate masks are connected to the same object while some possible candidates are neglected. To deal with this issue, as shown in the right bottom part of Figure~\ref{fig:motivation}, we impose the diversity loss on candidate object kernels to make them dissimilar between each other. This encourages generating diverse candidate masks, such that the objects in video can be all covered as possible. Here, candidate object kernels are learned from visual-text features by Cross-Modal Alignment module.

The main contributions of this work are highlighted as:
\begin{itemize}
	\item We propose an end-to-end RVOS framework completely built upon transformers, \ie, Fully Transformer-Equipped Architecture, termed FTEA. 
	
	\item We develop the stacked attention mechanism to model object-level spatial context, and the stacked FFN to reduce model parameters when obtaining independent candidate object feature maps by group linear mapping.
	
	\item We introduce the diversity loss on candidate object kernels to diversify candidate object masks.
	
	\item Extensive experiments were carried out on three benchmarks to validate the effectiveness of our FTEA approach.
	
\end{itemize}

The remainder of this paper is structured as follows. Section~\ref{related} reviews some closely related works and Section~\ref{method} introduces our FTEA framework for RVOS task. Then, we report experimental results on three benchmarks to show the superiority of our method in Section~\ref{test} followed by the discussion in Section~\ref{discuss}. Finally, this work is concluded in Section~\ref{conclusion}.
%

\section{Related Work}
\label{related}
\subsection{Referring Video Object Segmentation}
Different from conventional VOS \cite{zhen-ipm2022-segmentation,hong-tip-2021-adaptivevos,li-arxiv2023-lsta} and semantic segmentation \cite{ou-ipm2022-scene}, RVOS aims to segment the pixel-level target object in video, given a natural language description, \ie, text query or referring expression. This desires visual-linguistic understanding, and establishes the relation between the referred object and its corresponding pixel region of video.

In RVOS, dynamic convolution \cite{gavrilyuk-cvpr2018-aavs} as a typical method, introduces the dynamic kernels conditioned on text features and convolves with visual features to obtain object masks. While it is simple and efficient, but fails to well capture the pixel-level spatial context, since convolution kernels heavily depend upon referring expressions. To overcome this drawback, Wang \etal~\cite{wang-aaai2020-cmdn} proposed Context Modulated Dynamic Networks (CMDA) to generate dynamic kernels by using both visual and linguistic features, such that the object discriminant ability can be enhanced. Besides, several works adopt the attention mechanism, which not only globally encodes video frames and referring expressions, but also models the fine-grained semantic relations, including word-word, pixel-pixel, and word-pixel relations. For example, Wang \etal~\cite{wang-iccv2019-acga} designed the vision-guided language attention to reduce the linguistic variation of text query and the language-guided vision attention to obtain the object region related to text query. Since vanilla attention mechanism does not encode pixel-wise object position \cite{vaswani-nips2017-attention}, Ning \etal~\cite{ning-ijcai2020-prpe} developed the Polar Relative Positional Encoding (PRPE) mechanism to represent object spatial relation described by text query. To characterize more accurate object appearance, Ye \etal~\cite{ye-pami2021-cmsa} employ cross-modal self-attention to capture both the pixel-level and the word-level relations of linguistic and visual features. To leverage temporal coherence among video frames and obtain consistent object masks, Seo \etal~\cite{seo-eccv2020-urvos} introduced memory attention to model pixel-wise correlation across frames. Furthermore, Hui \etal~\cite{hui-cvpr2021-cstm} utilized a cross-modal attention module to enable the interaction between linguistic and visual features at both frame-level and video clip-level to obtain more robust object representations. In addition to vanilla dot product attention, Liu \etal~\cite{liu-pami2021-cmpc} proposed a temporal graph reasoning strategy on top of cross-modal attention maps to highlight the target object. Furthermore, some works take advantage of both dynamic convolution and attention mechanism, \eg, Botach \etal~ \cite{botach-cvpr2022-mttr} adopted the popular Transformer blocks \cite{vaswani-nips2017-attention} with attention to obtain more informative object-level features while using dynamic convolution to produce object masks. Following this work, Wu \etal~\cite{wu-cvpr2022-referformer} treated the language as queries, which are transformed into dynamic kernels for yielding more robust object features, but it is computationally expensive. Differently, Bellver \etal~\cite{bellver-arxiv2020-refvos} used frame-level features to obtain final object masks for better efficiency, but failed to capture temporal relations across frames, leading to inconsistent object prediction. Thereafter, Kazakos \etal~\cite{kazakos-naaclw2021-synthref} tried to generate synthetic referring expressions to improve the model generalization ability given different text queries. Instead of using convolution or transformer, Mcintosh \etal~\cite{mcintosh-cvpr2020-vtcr} encoded both the video and the text input in the form of capsules \cite{hinton-iclr2018-capsule} to obtain more effective cross-modal object representations.

Apart from the above end-to-end approaches, other RVOS methods \cite{yang-arxiv2020-aamn,liang-arxiv2021-rethinking,ding-arxiv2021-pmin} adopt the multi-stage pipeline, \ie, ensemble learning. For example, Chen \etal~\cite{chen-acmmm2021-ccan} estimated the initial object location using object proposals, which are derived from offline-trained instance segmentation model \cite{he-iccv2017-maskrcnn}; Liang \etal~\cite{liang-arxiv2021-rethinking} introduced a more complex but powerful pipeline, which is composed of two instance segmentation models \cite{chen-cvpr2019-htc, tian-eccv2020-condinst} and one VOS model \cite{yang-eccv2020-cfbi} to produce more accurate masks. Besides, Ding \etal~\cite{ding-arxiv2021-pmin} proposed the referring expression comprehension and segmentation model \cite{luo-cvpr2020-mcn} and one VOS model \cite{yang-eccv2020-cfbi} to obtain pixel-level object regions. In addition, Yang \etal~\cite{yang-tip-2022-textvos} tried to align the video content with the textual query in a fine-grained manner to alleviate the semantic asymmetry problem, while Yang \etal~\cite{yang-tip-2022-transrvos} conducted intra-modal and inter-modal joint learning for video referring segmentation without the aid of object detection or category-specific pixel labeling. 

\subsection{Transformers}
Transformer was proposed by Vaswani \etal~\cite{vaswani-nips2017-attention} as an attention-based building block for the sequence-to-sequence machine translation task. In recent, transformer has gained much popularity in natural language processing, and also been successfully applied to computer vision, such as object detection \cite{carion-eccv2020-detr,zhu-iclr-2021-ddetr}, visual tracking \cite{yan-iccv2021-lstt,chen-cvpr2021-tt}, semantic segmentation \cite{cheng-nips2021-maskformer,bao-iclr2022-beit}, and video captioning \cite{li-ipm2023-tftd}. Here we discuss several typical transformers.

Vision Transformer (ViT) \cite{dosovitskiy-iclr2021-vit} shows CNNs are unnecessary for image classification and a pure Transformer applied to sequences of image patches works as well. To handle large variations of visual entities and the high resolution of image pixels, Liu \etal~\cite{liu-iccv2021-swin} presented a hierarchical Transformer with Shifted windows (Swin) for attention mechanism, and Swin Transformer has been adapted to video recognition \cite{liu-arxiv2021-vswin}. Transformers not only can be used as vision classifier, but also can be use in object detection. For example, DETR (Detection with Transformers) \cite{carion-eccv2020-detr} introduces transformers into object detection, by employing a set of object queries as candidates, which are fed into Transformer as the decoder to obtain a final set of detection predictions. In addition, VisTR (Video instance segmentation with Transformers) \cite{wang-cpvr2021-vistr} extends the DETR framework to video instance segmentation (VIS) \cite{yang-iccv2019-vis} task and tackles VIS in an end-to-end manner with parallel sequence decoding. 

Recently, transformers have exhibited its strong power in semantic segmentation. For instance,  For instance, Maskformer \cite{cheng-nips2021-maskformer} is the transformer based model which addresses both semantic and panoptic segmentation tasks using mask classification paradigm. This paradigm disentangles the image partitioning and classification aspects of segmentation, which leverages both semantic-level and instance-level characteristics. Our model adopts the similar paradigm but uses different architecture that employs stacked transformers with an imposed diversity constraint. Inspired by this, we also use mask classification to distinguish objects in video as candidates, and utilize the associated referring scores conditioned on text query to predict the target object. Besides, previous transformer-based models still use convolution building block to decode pixel-wise features into masks, failing to fully exploit the transformer advantage. This motivates us to further develop a transformer based mask decoder for deriving more discriminant object-wise features.
%

\begin{figure*}
	\centering
	\includegraphics[width=0.95\linewidth]{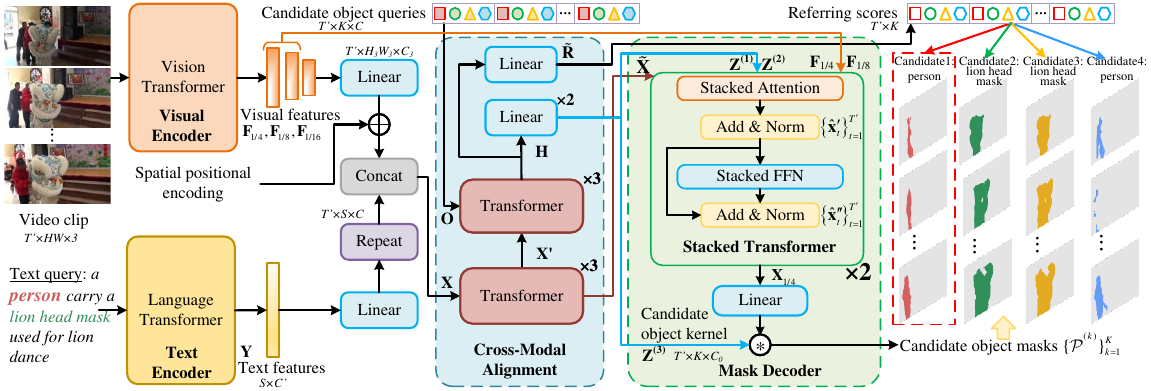}
	\caption{Overview of our Fully Transformer-Equipped Architecture (FTEA) framework for RVOS task.}
	\label{fig:framework}
\end{figure*}

\section{Our Method}
\label{method}

The methodology section first covers the problem definition with the model architecture overview, and then elaborates the details of each component in the FTEA framework, followed by the loss function descriptions in the end.

\subsection{Problem Definition}
For a video sequence with $n$ frames, each frame is accompanied by a text query $\mathcal{E}=\{w_s\}_{s=1}^{S}$ with $S$ words, the RVOS task aims to produce a series of frame-wise binary segmentation masks of the target object referred in the text query. As a common practice, the frames are randomly sampled from one video sequence to form a clip in each epoch during training, and the number of selected frames is called \textit{temporal length} or \textit{window size}. Thus, for a video clip $\mathcal{V} = \{\mathbf{I}_t\in\mathbb{R}^{ H\times W\times 3}|t=1,2, \ldots,T\} \subseteq \mathcal{V}$ with $T$ frames, where $\mathbf{I}_t$ denotes the $i$-th RGB frame with the height $H$, the width $W$, and three channels, the goal is to obtain corresponding $T$ binary segmentation masks $\hat{\mathcal{P}}=\{\mathbf{\hat{P}}_t\in \{0,1\}^{ H\times W}|t=1,2, \ldots,T\}$, where $\mathbf{\hat{P}}_t$ denotes the segmentation mask of $i$-th frame.

\subsection{The FTEA Framework}
Previous end-to-end approaches \cite{gavrilyuk-cvpr2018-aavs,seo-eccv2020-urvos,ye-pami2021-cmsa, chen-acmmm2021-ccan} commonly formulate RVOS as a per-pixel binary classification problem, applying a classification loss to each output pixel. However, when handling the text query involving multiple objects per-pixel binary classification cannot well capture multiple objects since the other objects are viewed as background except for the referred one. Hence, we introduce an alternative paradigm called mask classification \cite{cheng-nips2021-maskformer}, which predicts a set of binary masks, and each mask is associated with a single class prediction. Throughout this paper, we regard all appearing objects in video as candidate objects and attempt to produce binary masks for them all. For each binary mask, we use referring score, \aka, confidence score, to indicate whether the object is referred in text query and visible in video.

Given a video clip $\mathcal{V}$, an RVOS model $\Phi(\cdot)$ will produce a set of mask sequences $\tilde{\mathcal{P}}^{(k)}$ for $K$ candidate objects and the corresponding set of referring scores $\tilde{\mathcal{R}}^{(k)}$ associated with each object mask, \ie, $\Phi(\mathcal{V}) \rightarrow \{ (\tilde{\mathcal{P}}^{(k)}, \tilde{\mathcal{R}}^{(k)} ) \}_{k=1}^K$. Here, $\tilde{\mathcal{P}}^{(k)}$ contains $T$ object masks $\tilde{\mathbf{P}}^{(k)}_{t}\in \{0,1\}^{H\times W}$, and $\tilde{\mathcal{R}}^{(k)}=\{\tilde{r}_{k,t} \}_{t=1}^{T}$ contains $T$ probability values $\tilde{r}_{k,t}$, \ie, confidence score, indicating whether the $k$-th candidate object is referred and visible in the $t$-th frame. Usually, the number of candidate objects $K$ is set to be much larger than the number of objects $N$ in video. Next, the mask sequence with the highest confidence score is selected to produce final masks of the referred object. 

In principle, we utilize transformer \cite{vaswani-nips2017-attention} as building block to construct our RVOS model, which is termed as Fully Transformer Equipped Architecture (FTEA), whose entire framework is depicted in Figure~\ref{fig:framework}. It contains four main components, \ie, Visual Encoder that extracts multi-scale spatiotemporal features, Text Encoder that extracts compact text features, Cross-Modal Alignment module that aligns visual and text features, and Mask Decoder that produces final prediction by decoding object-level features into masks using newly developed Stacked Transformer. 

The working mechanism of FTEA is briefly described as: Firstly, video clip and text query are respectively fed into Visual Encoder and Text Encoder for obtaining visual features and text features, which are then concatenated into visual-textual features. Next, the transformer based Cross-Modal Alignment module is used to capture the frame-wise global visual-linguistic context, including pixel-pixel spatial relation, pixel-word semantic relation, and word-word semantic relation, resulting in the visual-text alignment features with candidate object kernels and referring scores. Then, the stacked transformer serves as the main component of Mask Decoder, which progressively learns object-level features that are decoded into $K$ mask sequences using candidate object kernels. Finally, the relevant mask sequences are selected for supervision during training, according to referring scores and mask quality in terms of Dice coefficient \cite{milletari-3dv2016-dice}, while the mask sequence with the highest referring score is picked as the final prediction during inference.

\subsection{Visual and Text Encoder}
For a video clip $\mathcal{V}=\{\mathbf{I}_t\in\mathbb{R}^{ H\times W\times 3}\}_{t=1}^{T}$, we use the transformer based visual encoder \cite{liu-arxiv2021-vswin} to generate pixel-level high, middle, and low-resolution appearance feature maps, \ie, $\mathbf{F}_{1/4}\in\mathbb{R}^{T\times H_1  W_1\times C_1}$, $\mathbf{F}_{1/8}\in\mathbb{R}^{T\times H_2 W_2 \times C_2}$, $\mathbf{F}_{1/16}\in\mathbb{R}^{T\times H_3 W_3\times C_3}$ , where $C_1=96, C_2=192, C_3=384$ are channel numbers and $H_1  W_1 = \frac{H}{4}\cdot \frac{W}{4}$, $H_2 W_2=\frac{H}{8}\cdot \frac{W}{8}$, $H_3 W_3=\frac{H}{16}\cdot \frac{W}{16}$ are the feature vector lengths of the down-sampled feature maps. For text query $\mathcal{E}=\{w_s\}_{s=1}^{S}$ with $S$ words, we adopt transformer based linguistic model RoBERTa (Robustly optimized BERT approach) \cite{liu-arxiv2019-roberta}, to extract the word-level text feature $\mathbf{Y}\in\mathbb{R}^{S\times C'}$ ($C'=768$). Then, the low-resolution appearance feature map $\mathbf{F}_{1/16}$ and text feature $\mathbf{Y}$ are linearly projected to the shared data space with dimension $C$ ($C=256$). Next, the projected low-resolution appearance feature map is flattened and concatenated with the projected text feature to produce the visual-text feature tensor $\mathbf{X}\in\mathbb{R}^{T\times( H_3 W_3 +S)\times C}$. 

\subsection{Cross-Modal Alignment}
\label{sec:cross_modal_align}
The cross-modal alignment module mainly consists of vanilla transformer encoders and transformer decoders. These transformer blocks adopt scale dot product attention and linear layers. The former attends the region of interest in each frame by encoding visual-text feature $\mathbf{X}$, while the latter learns a set of candidate object queries (\ie, object feature vectors) using the output features of the former, which is similar to the object queries in DETR \cite{carion-eccv2020-detr}.  

For transformer encoders (three here), each one has a standard architecture and consists of multi-head self-attention module and feed forward network (FFN). Following \cite{carion-eccv2020-detr}, we add the fixed sine spatial positional encoding to the appearance feature map of each frame while no positional encoding is used for text features. For memory efficiency, we divide the temporal dimension into $T'$ groups for parallel computing. Its output is the visual-text alignment feature $\mathbf{X}'\in\mathbb{R}^{T\times (H_3 W_3 +S)\times C}$.

For transformer decoders (three here), we introduce $K$ candidate object queries to represent candidate objects of each frame, as in \cite{wang-cpvr2021-vistr}. Specifically, we define a tensor $\mathbf{O}\in \mathbb{R}^{T\times K \times C}$ to accommodate candidate object queries. The queries share the weights across frames and learn to attend to the same candidate object in video. After that, we feed the candidate object query $\mathbf{O}$ with visual-text alignment feature $\mathbf{X}'$ to the vanilla transformer decoder, in order to globally reason about all candidate objects together and thus model the semantic relations of object-word and object-pixel pairs. Then, we compute the attentive candidate object query tensor $\mathbf{O}'\in \mathbb{R}^{T\times K \times C}$, \ie,  
\begin{equation}
	\label{eq:query_attn}
	\mathbf{O}' = \text{LN}(\text{mh-attn}(\mathbf{O} \mathbf{W}'_{query}, \mathbf{X}' \mathbf{W}'_{key}, \mathbf{X}' \mathbf{W}'_{value}) + \mathbf{O}),
\end{equation}
where $\{\mathbf{W}'_{query}, \mathbf{W}'_{key}, \mathbf{W}'_{value} \} \in \mathbb{R}^{C\times C}$ are learnable parameter matrices, and $\mathbf{O}'$ retrieves relevant object-level semantics from visual-text alignment feature $\mathbf{X}'$ by exploring their pair-wise relations. Next, the attentive candidate object queries are fed to one FFN to obtain the first hidden feature $\mathbf{O}_1\in \mathbb{R}^{T\times K \times C}$, which goes through the subsequent two transformer decoders with the same visual-text alignment feature $\mathbf{X}'$ to obtain finer hidden features $\{\mathbf{O}_2, \mathbf{O}_3\} \in \mathbb{R}^{T\times K \times C}$. By concatenating these hidden features, it outputs hidden feature of candidate object as $\mathbf{H}=[\mathbf{O}_1; \mathbf{O}_2; \mathbf{O}_3 ]\in\mathbb{R}^{3\times T\times K\times C}$, where $[~;~]$ denotes the concatenation. This results in coarse-to-fine hidden features for learning finer details of objects.

To incorporate object-level spatial location with pixel-level visual feature, inspired by \cite{gavrilyuk-cvpr2018-aavs,wang-aaai2020-cmdn}, we propose to use dynamic convolution strategy to produce the mask sequence for each candidate object. Particularly, two linear layers with ReLU in between are operated on the hidden feature $\mathbf{H}$ to derive several dynamic kernels $\{\mathbf{Z}^{(1)}, \mathbf{Z}^{(2)}, \mathbf{Z}^{(3)}\} \in\mathbb{R}^{T\times K\times C_0}$, where $C_0=8$. Since the hidden feature is learned from attentive candidate object query $\mathbf{O}'$ to finely represent candidate objects, dynamic kernels are also called candidate object kernels. These dynamic kernels are then fed into the following mask decoder, for progressively decoding pixel-wise object features into $K$ candidate object mask sequences.

To match the mask with the referred object, we introduce a referring score $\tilde{r}_{k,t}$ for each candidate object mask, which is a probability produced by passing hidden feature of candidate object $\mathbf{H}$ to a linear layer with sigmoid function. Thus, we obtain referring score matrix $\tilde{\mathbf{R}}\in \mathbb{R}^{T\times K}$, whose entries are between 0 and 1. 

\subsection{Mask Decoder}
Previous works \cite{seo-eccv2020-urvos,botach-cvpr2022-mttr} usually adopt convolution based Feature Pyramid Network (FPN) \cite{lin-cvpr2017-fpn} to decode pixel-wise feature maps into masks. However, the convolution only aggregates pixel-level features from the local region of feature map and these features cannot well reflect rich semantics, \eg, car, person, and bicycle all appear in one image. This desires modeling the object-level appearance pattern that embeds latent semantics inherently. Hence, to model object-level spatial relations for candidate objects, we introduce a progressive object learning scheme by designing Stacked Transformer, which consists of two modules, \ie, \textit{Stacked Attention} and \textit{Stacked FFN}. They \textit{stack} the object-level semantic features of candidate objects sequentially along the channel dimension in order to avoid mask confusion. 

In detail, the candidate object kernels from cross-modal alignment module are fed into the Stacked Transformer to incorporate object-level context. Similar to FPN, Stacked Transformer utilizes the rich appearance cues by fusing high-resolution feature maps derived from visual encoder across frames in video. Here, we adopt two cascaded Stacked Transformers to build our mask decoder. To produce pixel-wise mask, we remove the text part of visual-text alignment feature $\mathbf{X}'\in\mathbb{R}^{T\times (H_3 W_3 +S)\times C}$ to obtain visual alignment feature $\tilde{\mathbf{X}}\in\mathbb{R}^{T\times H_3  W_3 \times C}$.

For the first Stacked Transformer, we take the visual alignment feature $\tilde{\mathbf{X}}$, middle-resolution appearance feature $\mathbf{F}_{1/8}\in\mathbb{R}^{T\times H_2 W_2 \times C_2}$, and the first candidate object kernel $\mathbf{Z}^{(1)}\in\mathbb{R}^{T\times K\times C_0}$ as input, and it outputs middle-resolution decoding feature $\mathbf{X}_{1/8}\in \mathbb{R}^{T\times H_2W_2 \times \alpha K}$, where $\alpha$ (set to 4) is the feature channel number of each candidate object. For memory efficiency, we split the temporal dimension into $T$ groups for parallel computing. Thus, we denote the visual alignment feature as $\tilde{\mathbf{X}}=\{\tilde{\mathbf{x}}_{t}\}_{t=1}^{T}$, middle-resolution appearance feature as $\mathbf{F}_{1/8}=\{\mathbf{f}_t \}_{t=1}^{T}$, and the first candidate object kernel as $\mathbf{Z}^{(1)}=\{ \mathbf{z}_{t} \}_{t=1}^{T}$, where $\tilde{\mathbf{x}}_{t}\in\mathbb{R}^{ H_3  W_3 \times C}$, $ \mathbf{f}_t \in\mathbb{R}^{H_2 W_2 \times C_2}$, $\mathbf{z}_{t}\in\mathbb{R}^{ K\times C_0} $. For the second Stacked Transformer, we employ middle-resolution decoding feature $\mathbf{X}_{1/8}$, high-resolution appearance feature map $\mathbf{F}_{1/4}$, and the second candidate object kernel $\mathbf{Z}^{(2)}$ as input to produce high-resolution decoding feature $\mathbf{X}_{1/4}\in\mathbb{R}^{T\times H_1  W_1\times \alpha' K}$, where $\alpha'$ is set to 2. 

\noindent\textbf{Stacked Attention}. We first feed each visual alignment feature and middle-resolution appearance feature into linear layers to produce a 3-tuple $(query, key, value)$ as:
\begin{equation}
	\mathbf{q}''_t = \mathbf{f}_{t} \mathbf{W}''_{query},\\
	\tilde{\mathbf{k}}_t = \tilde{\mathbf{x}}_t \mathbf{W}''_{key},\\
	\tilde{\mathbf{v}}_t = \tilde{\mathbf{x}}_t \mathbf{W}''_{value},
\end{equation}
where $\mathbf{W}''_{query}\in\mathbb{R}^{C_2\times K}$, $\mathbf{W}''_{key}\in\mathbb{R}^{C\times K}$ and $\mathbf{W}''_{value}\in\mathbb{R}^{C\times \alpha K}$ are learnable parameters; $\mathbf{q}''_t \in\mathbb{R}^{H_2  W_2 \times K}$, $\tilde{\mathbf{k}}_t\in\mathbb{R}^{H_3  W_3\times K}$ and $\tilde{\mathbf{v}}_t \in\mathbb{R}^{ H_3 W_3 \times \alpha  K}$ denote the initial query, key and value in attention mechanism. 

Then, we incorporate object-level semantics into pixel-wise features using candidate object kernel, leading to candidate object weight matrix $\mathbf{m}_t\in \mathbb{R}^{H_2 W_2\times K}$ as
\begin{equation}
	\mathbf{m}_t = \sigma(\text{Upsample1}(\mathbf{z}_t * ( \tilde{\mathbf{x}}_t \mathbf{W}_0) )) ,
\end{equation}
where $ \mathbf{W}_0\in \mathbb{R}^{C\times C_0}$ is a learnable parameter matrix; $\text{Upsample1}(\cdot)$ denotes bilinear interpolation to upsample the resolution of input feature map from $H_3\times W_3$ to $H_2\times W_2$; $\sigma(\cdot)$ denotes a sigmoid function; $'\ast'$  denotes a $1\times 1$ dynamic convolution operation, which can be viewed as a linear projection using candidate object kernel $\mathbf{z}_{t}$ as the projection matrix. We apply an element-wise product $\odot$ between query $\mathbf{q}''_t$ and weight matrix $\mathbf{m}_t$ to enhance the object awareness of the query as 
\begin{equation}
	\tilde{\mathbf{q}}_t = \mathbf{q}''_t \odot \mathbf{m}_t \in\mathbb{R}^{H_2 W_2\times K},
\end{equation}
which has $K$ feature channels and each channel reflects the visual appearance of each candidate object. These channels are stacked together to shape a feature bank in order. Thus, we can utilize the object-level spatial context to capture finer appearance cues. Next, the query $\tilde{\mathbf{q}}_t$ interacts with the initial key and value derived from visual alignment feature, using cross attention, \ie,
\begin{equation}
	\label{eq:stacked_atention}
	\text{Att}(\tilde{\mathbf{q}}_t, \tilde{\mathbf{k}}_t ,\tilde{\mathbf{v}}_t)  = \text{Softmax}(\frac{\tilde{\mathbf{q}} _t\tilde{\mathbf{k}}_t^T}{\sqrt{K}})\tilde{\mathbf{v}}_t,
\end{equation}
where $\text{Softmax}(\cdot)$ denotes softmax function and the output is the initial attentive object feature $\mathbf{\hat{x}}_t\in\mathbb{R}^{ H_2 W_2\times \alpha  K}$. 

Essentially, the query $\tilde{\mathbf{q}}_t$ encodes fine details (edge and texture) and coarse object-level spatial location, the key $\tilde{\mathbf{k}}_t$ encodes the coarse appearance feature of each candidate object and finer object-level location, the value $\tilde{\mathbf{v}}_t$ enriches object-level semantics. Using the dot product attention in Eq.~(\ref{eq:stacked_atention}), object-level appearance cues can be retrieved from visual alignment feature and fused with finer details, resulting in more accurate object mask sequence for each candidate object. In addition, a residual connection is used to produce attentive object feature $\mathbf{\hat{x}}_t'$ by
\begin{equation}
	\label{eq:attention_residual}
	\mathbf{\hat{x}}' _t=\mathbf{\hat{x}}_t+ \text{Upsample1}(\tilde{\mathbf{v}}_t) \in\mathbb{R}^{H_2 W_2\times \alpha K}.
\end{equation}
This takes into account the upsampled value $\tilde{\mathbf{v}}_t$, which enriches visual semantics of candidate objects.

\noindent\textbf{Stacked FFN.} To preserve the channel order of object-level features, we propose a group-wise multi-layer perceptron as an implementation of Feed Forward Network \cite{vaswani-nips2017-attention}, termed Stacked FFN (SFFN). Since each group of perceptron reveals the visual semantics of one candidate object, the obtained object-level features are sequentially stacked in a feature pool. Specifically, we divide attentive object feature $\mathbf{\hat{x}}_t'$ into $K$-groups along the channel dimension, and each group is assigned with $\alpha$ channels. 

Besides, we employ group-wise convolution to design two linear layers to capture object-level appearance independently. Meanwhile, layer normalization \cite{ba-arxiv2016-layernorm} and residual connection are used to produce the feature $\mathbf{\hat{x}}''_t$, \ie,
\begin{equation}
	\label{eq:sffn}
	\mathbf{\hat{x}}''_t= \text{LN}(\text{SFFN}(\text{LN}( \mathbf{\hat{x}}_t )) + \mathbf{\hat{x}}_t) \in\mathbb{R}^{H_2 W_2\times \alpha K},
\end{equation}
which acts as the middle-resolution decoding feature of the first Stacked Transformer for each frame. Then, we concatenate corresponding $T'$ output features to obtain $\mathbf{X}_{1/8}\in\mathbb{R}^{T\times H_2 W_2 \times \alpha K}$. Similarly, we can obtain $\mathbf{X}_{1/4}$ using the second Stacked Transformer.

After passing through the above two modules, we apply $1\times 1$ dynamic convolution using the third candidate object kernel $\mathbf{Z}^{(3)}$ on the linearly projected high-resolution decoding feature $\mathbf{X}_{1/4}$ to obtain $K$ sets of candidate object mask sequences, \ie, 
\begin{equation}
	\label{eq:last_dynamic_conv}
	\{ \tilde{\mathcal{P}}^{(k)} \}_{k=1}^K =\sigma(\text{Upsample2}(\mathbf{Z}^{(3)} * (\mathbf{X}_{1/4} 
	\mathbf{W}_{\text{proj}}) )),
\end{equation}
where $\mathbf{W}_{\text{proj}}\in\mathbb{R}^{\alpha'  K \times C_0}$ is a learnable parameter; $\text{Upsample2}(\cdot)$ denotes bilinear interpolation to upsample the resolution of input feature map from $H_1\times W_1$ to $H\times W$; $\tilde{\mathcal{P}}^{(k)}= \{ \tilde{\mathbf{P}}_{k,t}  \in \mathbb{R}^{ H\times W} \}_{t=1}^{T}$ contains $K$ candidate object masks for each frame. Besides, we convert the referring score matrix $\tilde{\mathbf{R}}\in [0,1]^{T\times K}$ in Sec.~\ref{sec:cross_modal_align} into referring score set $ \{ \tilde{\mathcal{R}}^{(k)} \}_{k=1}^{K}$, where $\tilde{\mathcal{R}}^{(k)}=\{r_{k,t} \}_{t=1}^{T}$. 

We associate masks and referring scores accordingly to form pair-wise predictions for the $k$-th candidate object as 
\begin{equation}
	\hat{\mathcal{Y}}^{(k)} = (  \tilde{\mathcal{P}}^{(k)}, \tilde{\mathcal{R}}^{(k)} ).
\end{equation}

\subsection{Loss Function}
We adopt the sum of Dice loss \cite{milletari-3dv2016-dice} and Focal loss \cite{lin-iccv2017-focal}, \ie, $\mathcal{L}_{\text{mask}}$, to supervise the mask prediction at frame-level, while using the binary cross entropy loss $\mathcal{L}_{\text{ref}}$ to supervise the referring score prediction. Details are below.

To select the best matched candidate object mask sequence, we generate the ground-truth referring score $r_{k',t}$, according to the ground-truth mask indicating the referred object in each frame. When the referred object is visible, its value is set to 1, otherwise 0. In some cases, there is more than one text query for a video, \ie, more than one referred objects. For the ground-truth referring score set $\{ \mathcal{R}^{(k')} \}_{k'=1}^K$, where $\mathcal{R}^{(k')}  = \{ r_{k',t}\in \{0,1\}\}_{t=1}^{T}$, we pad it with $\emptyset$ to fill in those missing slots. The symbol $\emptyset$ denotes that the ground-truth mask is unavailable for corresponding candidate object. Similarly, the ground-truth mask sequence set is denoted as $\{ \mathcal{P}^{(k')} \}_{k'=1}^K$, where $\mathcal{P}^{(k')}=\{ \mathbf{P}_{k',t}\in \{0,1\}^{H\times W}\}_{t=1}^{T}$. Then, the ground-truth sequence tuple for $k'$-th candidate object is 
\begin{equation}
	\mathcal{Y}^{(k')} = ( \mathcal{P}^{(k')}, \mathcal{R}^{(k')} ).
\end{equation}

To find a matching between ground-truth mask sequences and candidate object mask sequences, we apply pair-wise matching cost function \cite{botach-cvpr2022-mttr} as  
\begin{equation}
	\label{eq:matching_cost}
	\begin{aligned}
		\mathcal{C}_{\text{match}}(\hat{\mathcal{Y}}^{(k)}, \mathcal{Y}^{(k')}) = \lambda_{\text{dice}} \mathcal{C}_{\text{dice}}( \tilde{\mathcal{P}}^{(k)}, \mathcal{P}^{(k')} ) \\
		+ \lambda_{\text{ref}} \mathcal{C}_{\text{ref}}( \tilde{\mathcal{R}}^{(k)},\mathcal{R}^{(k')}),
	\end{aligned}
\end{equation}
where $\lambda_{\text{dice}}>0$ and $\lambda_{\text{ref}}>0$ are hyper-parameters, the first term $ \mathcal{C}_{\text{dice}}(\cdot)$ supervises the $k$-th candidate mask sequence using the $k'$-th ground-truth mask sequence by averaging negative Dice Coefficients \cite{milletari-3dv2016-dice} of each corresponding mask pair per time instance, and the second term $\mathcal{C}_{\text{ref}}(\cdot)$ supervises the predicted referring score using the corresponding ground-truth sequence as
\begin{equation}
	\mathcal{C}_{\text{ref}}( \tilde{\mathcal{R}}^{(k)},\mathcal{R}^{(k')}) = - \frac{1}{T}\sum_{t=1}^{T} r_{k',t} \cdot \tilde{r}_{k,t}.
\end{equation}

According to the cost function in Eq.~(\ref{eq:matching_cost}), we can find the optimal assignment with the minimum cost by Hungarian algorithm \cite{kuhn-nrlq1955-hungarian}, and calculate loss function only for the candidate objects assigned with nonempty ground truth. For $N$ referred objects, we define the new index of the candidate object as $k''=\delta(k)\in \{1,\ldots, N\}$, where $\delta(\cdot)$ denotes the optimal Hungarian assignment.

Besides, we adopt the Dice loss \cite{milletari-3dv2016-dice} and Focal loss \cite{lin-iccv2017-focal} to supervise the mask prediction as
\begin{equation}
	\begin{aligned}
		\mathcal{L}_{\text{mask}}(\tilde{\mathbf{P}}_{k'',t},\mathbf{P}_{k',t}) =  \sum_{t=1}^{T} (  \lambda_{\text{dice}}\mathcal{L}_{\text{dice}}( \tilde{\mathbf{P}}_{k'',t},\mathbf{P}_{k',t} )\\
		+ \lambda_{\text{focal}}\mathcal{L}_{\text{focal}}( \tilde{\mathbf{P}}_{k'',t},\mathbf{P}_{k',t} ) ),
	\end{aligned}
\end{equation}
where $\lambda_{\text{focal}}>0$ is a hyper-parameter, $\mathcal{L}_{\text{dice}}(\cdot)$ denotes the Dice loss which calculates negative Dice Coefficient for every mask pair, and $\mathcal{L}_{\text{focal}}(\cdot)$ is a cross entropy loss which can alleviate the class imbalance problem at pixel-level by focusing on hard pixels.

Besides, we utilize the binary cross entropy loss to supervise the referring score prediction, \ie,
\begin{equation}
	\mathcal{L}_{\text{ref}}(\tilde{r}_{k'',t},r_{k',t}) = - \lambda_{\text{ref}} \sum_{t=1}^{T} r_{k',t} \log(\tilde{r}_{k'',t}).
\end{equation}
Following \cite{carion-eccv2020-detr, botach-cvpr2022-mttr}, we downweight the loss values of negative (``unreferred") candidate objects by a factor of 10 to account for class imbalance. 

In fact, Dice loss and Focal loss cannot well model object-level relations. However, our model adopts the mask classification paradigm, which produces a number of candidate object mask sequences, where the matched one is selected as final prediction. Naturally, we hope each produced mask can depict a unique candidate object and different candidate objects can be characterized with distinct masks. Hence, to capture as many objects as possible in video, we introduce the diversity loss on candidate object kernels for yielding diverse candidate object masks, \ie,
\begin{small}
	\begin{equation}
		\mathcal{L}_{\text{div}}(\mathbf{Z}^{(1)}, \mathbf{Z}^{(2)}, \mathbf{Z}^{(3)}) = \sum_{t=1}^{T}\sum_{j=1}^{3} ||	\mathbf{Z}_{j,t} \mathbf{W}_{\text{div}} \mathbf{Z}_{j,t}^T -\mathbf{I}||_F +  || \mathbf{W}_{\text{div}}||_1,
	\end{equation}
\end{small}%
where $||\cdot||_1$ denotes $\ell_1$-norm, $||\cdot ||_F$ denotes Frobenius norm, $\mathbf{W}_{\text{div}}\in\mathbb{R}^{C_0\times C_0}$ is a learnable parameter, $\mathbf{I}\in\mathbb{R}^{C_0\times C_0}$ is an identity matrix, and $\mathbf{Z}_{j,t}\in\mathbb{R}^{K\times C_0}$ denotes the $j$-th candidate object kernels of the $t$-th frame, \ie, the $C_0$-dimensional feature representation of $K$ candidate objects.

Specifically, we first do a linear mapping with learnable matrix $\mathbf{W}_{\text{div}}$ on candidate object kernel $\mathbf{Z}_{j,t}$ to model the pair-wise similarity of candidate objects. It is followed by a matrix multiplication between the projected object kernel and the original one, resulting in pair-wise similarity scores. In addition, we subtract an identity matrix for ignoring the similarity of candidate objects themselves.  In addition, we apply $\ell_1$-norm on the weight matrix $\mathbf{W}_{\text{div}}$ to prevent over-fitting. By minimizing the normalized similarity score, the model is forced to learn different object-level features to produce diverse candidate object kernels. In this way, diverse masks can be generated by utilizing the dynamic convolution with diverse candidate object kernels, which further helps to obtain accurate masks of the referred object. 

Therefore, the final objective function of our FTEA model is formulated as
\begin{equation}
	\label{eq:loss_final}
	\mathcal{L} =  \mathcal{L}_{\text{mask}} + \mathcal{L}_{\text{ref}} + \lambda_{\text{div}} \mathcal{L}_{\text{div}},
\end{equation}
where $\lambda_{\text{div}} >0$ is a hyper-parameter for balancing the contribution of the diversity term.  

\section{Experiments}
\label{test}
All the experiments were conducted on a machine with two NVIDIA TITAN RTX Graphic Cards for training and inference, and our model was compiled using PyTorch 1.10, Python 3.9 and CUDA 11.1.

\begin{table}[!h]
	\centering
	\caption{Statistics of the datasets.}
	\label{table:statistics}
	\setlength{\tabcolsep}{1.4mm}{
		\begin{tabular}{l|ccc|cc}
			\toprule[0.75pt]
			\multirow{1}{*}{Dataset} &\multirow{1}{*}{Train}&\multirow{1}{*}{Val} &\multirow{1}{*}{Test}& \multirow{1}{*}{Class} & \multirow{1}{*}{Sentence} \\
			\midrule[0.5pt]	
			A2D Sentences\cite{xu-cvpr2015-a2d}	         	&3036	&-		&746	&8		&6655	\\
			J-HMDB Sentences \cite{gavrilyuk-cvpr2018-aavs}		&-		&-		&928	&21		&928	\\
			Ref-YouTube-VOS\cite{seo-eccv2020-urvos}		&3471	&202	&305	&94		&27899  \\
			\toprule[0.75pt]
		\end{tabular}
	}
\end{table}

\subsection{Data Sets}
We conduct extensive experiments on three datasets: A2D Sentences \cite{gavrilyuk-cvpr2018-aavs}, J-HMDB Sentences \cite{gavrilyuk-cvpr2018-aavs}, and Ref-YouTube-VOS \cite{seo-eccv2020-urvos}.  Some statistics of the them are listed in Table~\ref{table:statistics} and details are shown below.

\noindent\textbf{A2D Sentences} \cite{gavrilyuk-cvpr2018-aavs} is extended from Actor-Action database \cite{xu-cvpr2015-a2d} by adding textual descriptions for each video. It contains 3782 videos annotated with 8 action classes and 6655 sentences in total. For each video, there are 3 to 5 frames annotated with pixel-wise segmentation masks. There are 3036 training videos and 746 test videos, respectively. 

\noindent\textbf{J-HMDB Sentences} \cite{gavrilyuk-cvpr2018-aavs}, extended from J-HMDB database \cite{jhuang-iccv2013-jhmdb}, contains 21 different actions, 928 videos and corresponding 928 sentences. For each video, there are frame-wise 2D articulated human puppet masks. All actors are humans and one natural language query is annotated to describe the performed action.

\noindent\textbf{Ref-YouTube-VOS} \cite{seo-eccv2020-urvos} is extended from YouTube-VOS database \cite{xu-arxiv2018-ytbvos}. It contains 3975 videos, 7451 objects and 27899 text expressions at both first-frame and full-video levels. The first-frame expressions only describe the target object in the first frame, while the full-video expressions describe it through the whole video. But only the subset with more challenging full-video expressions are publicly available and is split into training set, validation set, and test set with 3471, 202, and 305 videos, respectively. Since there are only ground-truth annotations for training while the test server is now closed, we upload our mask predictions of validation set to the competition server\footnote{https://competitions.codalab.org/competitions/29139} to derive results.

\subsection{Evaluation Metrics}
Following previous works \cite{botach-cvpr2022-mttr,gavrilyuk-cvpr2018-aavs}, we train our model on A2D training set while evaluating the segmentation performance on A2D test set and the entire J-HMDB Sentences data. Meanwhile, we adopt \textit{Overall IoU} (Intersection over Union), \textit{Mean IoU}, \textit{Precision@$\zeta$} and \textit{mAP} (mean Average Precision) as the pixel-wise evaluation criteria. 

Among them, \textit{Overall IoU} denotes the ratio between the total intersection and the total union area over all test samples; \textit{Mean IoU} denotes the average value of IoU over all test samples; \textit{Precision@$\zeta$} is the percentage of test samples whose IoU scores are higher than a threshold $\zeta\in [0.5, 0.6, 0.7, 0.8, 0.9]$; \textit{mAP} calculates the mean precision averaged over a group of IoU thresholds [0.5:0.05:0.95] by using the API\footnote{https://github.com/cocodataset/cocoapi} implementation of object detection benchmark Microsoft COCO \cite{lin-eccv2014-mscoco}. 

For Ref-YouTube-VOS, we follow \cite{seo-eccv2020-urvos} to adopt several evaluation metrics, including \textit{Region Similarity} ($\mathcal{J}$), \textit{Contour Accuracy} ($\mathcal{F}$), and their average score ($\mathcal{J}$\&$\mathcal{F}$). Here, $\mathcal{J}$ denotes the average value of IoU scores over all test samples, and $\mathcal{F}$ calculates the average F1 scores of the contour points over all test samples.

\subsection{Implementation Details}
In the default configuration, we set the temporal length $T$ (\aka, window size or frame number) of one clip to 8 and each video is divided into many equal-length clips. Note that the above data sets have provided sampled frames in each video and we just use them as is; if the ultimately remaining frames are less than the temporal length, we just use them all. For A2D Sentences and Ref-YouTube-VOS, the batch size is respectively set to 4 and 2, on 2 TITAN RTX 24GB GPUs. Due to limited GPU resources, both the batch size and the temporal length have been set to the possible maximal. Actually, the segmentation performance can be further boosted if the temporal length increases as indicated in \cite{botach-cvpr2022-mttr}.

We adopt the tiny version of the Video Swin Transformer \cite{liu-arxiv2021-vswin} pretrained on Kinetics-400 \cite{kay-arxiv2017-kinetics} as visual encoder to extract pixel-level features for input video clips. Following \cite{botach-cvpr2022-mttr}, we use the last stage features from the visual encoder as the input of cross-modal alignment module, and there are three spatial strides (\aka, down-sampling rates) 4, 8, and 16, for obtaining the features. To obtain per-frame features, the single temporal down-sampling layer is abandoned from visual encoder, by setting the kernel and stride of its 3D convolution to size $1\times 4\times 4$. Besides, we adopt the base version of RoBERTa (Robustly optimized BERT approach)\cite{liu-arxiv2019-roberta} implemented by Hugging Face \cite{wolf-2020emnlp-huggingface} as text encoder to extract word-level features, and this encoder is a transformer-based language representation model. The weights of cross modal alignment module are randomly initialized with Xavier initialization. Additive dropout of 0.1 is applied after after every multi-head attention and FFN before layer normalization in the cross modal alignment module. For the spatial positional encoding in the cross modal alignment model, we adopt a 2D case \cite{parmar-icml2018-pe-2d} instead of the original one from transformer \cite{vaswani-nips2017-attention}. In addition, we set the number of candidate objects to $K=50$, which is much larger than the actual number of objects in one video.

\textbf{Training}. For A2D Sentences, we feed the model $T$ frames with the annotated target frame in the middle, and each frame is downsampled to $320\times 576$. For Ref-YouTube-VOS, we feed the model $T$ consecutive annotated frames. Each frame is resized to $360\times 640$. The loss hyper-parameters are empirically set as: $\lambda_{\text{dice}}=5$,  $\lambda_{\text{ref}}=5$, $\lambda_{\text{focal}}=2$, $\lambda_{\text{div}}=0.07$. We utilize AdamW (Adam with decoupled Weight decay) \cite{loshchilov-iclr2019-adamw} as the optimizer with weight decay set to $10^{-4}$. We also apply gradient clipping with a maximal gradient norm of 0.1. To improve position awareness, we randomly flip the input frames horizontally and swap direction-related word in the corresponding text queries accordingly. A learning rate $\eta$ of $10^{-4}$ is used for the cross alignment module and mask decoder, while $5\times 10^{-5}$ for visual encoder. The parameters of text encoder are kept frozen for efficiency. We train our model on A2D Sentences for 70 epochs with a learning rate drop by a factor of 2.5 after the first 50 epochs. For Ref-YouTube-VOS, the model is trained for 30 epochs with a learning rate drop by a factor of 2.5 after the first 20 epochs. Note that we do not utilize the time-consuming pretraining process on static images, such as Microsoft COCO database \cite{lin-eccv2014-mscoco}, which could further boost the segmentation performance. To avoid over-lengthy training, we adopt early stopping strategy by setting the maximum epoch to 70 and 30 for A2D Sentences and Ref-YouTube-VOS, respectively.

\textbf{Inference}. For A2D Sentences and J-HMDB Sentences, each frame is resized so that the short side has at least 320 pixels and the long side has at most 576 pixels; for Ref-YouTube-VOS, we apply the same resize configuration as in training. To obtain the mask sequence of target object, it requires to identify the index of the highest referring score summation, \ie, $k^\ast =\mathop{\arg\max_{k=1,\ldots,K}} \sum_{t=1}^{T} \tilde{r}_{k,t}$, and then retrieve the corresponding mask sequence.

\begin{table*}[!t]
	\centering
	\caption{Comparison with SOTA methods on A2D Sentences and J-HMDB Sentences. DC:Dynamic Convolution; Att:Attention.}
	\label{table:a2d_jhmdb}
	\small
	\setlength{\tabcolsep}{0.8mm}{
		\begin{tabular}{lccc|cccccccc|cccccccc}
			\toprule[0.75pt]
			\multirow{3}{*}{Method} & \multicolumn{1}{l}{\multirow{3}{*}{Year}} & \multicolumn{1}{l}{\multirow{3}{*}{DC}} & \multicolumn{1}{l|}{\multirow{3}{*}{Att}} & \multicolumn{8}{c|}{A2D Sentences} & \multicolumn{8}{c}{J-HMDB Sentences} \\ \cline{5-20} 
			& \multicolumn{1}{l}{} & \multicolumn{1}{l}{} & \multicolumn{1}{l|}{} & \multicolumn{5}{c|}{Precision} & \multicolumn{1}{c|}{mAP} & \multicolumn{2}{c|}{IoU} & \multicolumn{5}{c|}{Precision} & \multicolumn{1}{c|}{mAP} & \multicolumn{2}{c}{IoU} \\ \cline{5-20} 
			& \multicolumn{1}{l}{} & \multicolumn{1}{l}{} & \multicolumn{1}{l|}{} & 0.5 & 0.6 & 0.7 & 0.8 & \multicolumn{1}{c|}{0.9} & \multicolumn{1}{c|}{0.5:0.95} & Overall & Mean & 0.5 & 0.6 & 0.7 & 0.8 & \multicolumn{1}{c|}{0.9} & \multicolumn{1}{c|}{0.5:0.95} & Overall & Mean \\ \midrule[0.5pt]
			AAVS\cite{gavrilyuk-cvpr2018-aavs} & 2018  & \checkmark  &  & 50.0 & 37.6 & 23.1 & 9.4 & \multicolumn{1}{c|}{0.4} & \multicolumn{1}{c|}{21.5} & 55.1 & 42.6  & 71.2 & 51.8 & 26.4 & 3.0 & \multicolumn{1}{c|}{0.0} & \multicolumn{1}{c|}{26.7} & 55.5 & 57.0 \\
			ACGA\cite{wang-iccv2019-acga} & 2019 &  & \checkmark & 55.7 & 45.9 & 31.9 & 16.0 &  \multicolumn{1}{c|}{2.0} & \multicolumn{1}{c|}{27.4} & 60.1 & 49.0 & 75.6 & 56.4 & 28.7 & 3.4 & \multicolumn{1}{c|}{0.0} & \multicolumn{1}{c|}{28.9} & 57.6 &  58.4 \\
			VT-Caps.\cite{mcintosh-cvpr2020-vtcr}& 2020 &  &  & 52.6 & 45.0 & 34.5 & 20.7   & \multicolumn{1}{c|}{3.6} & \multicolumn{1}{c|}{30.3} & 56.8 & 46.0 & 67.7 & 51.3 & 28.3 & 5.1 & \multicolumn{1}{c|}{0.0} & \multicolumn{1}{c|}{26.1} & 55.5 & 57.0 \\
			PRPE\cite{ning-ijcai2020-prpe} & 2020  &  & \checkmark & 63.4 & 57.9 & 48.3 & 32.2  & \multicolumn{1}{c|}{8.3} & \multicolumn{1}{c|}{38.8} & 66.1 & 52.9 & 69.1 & 57.2 & 31.9 & 6.0 & \multicolumn{1}{c|}{\textbf{0.1}} & \multicolumn{1}{c|}{29.4} & - & - \\
			CMDy\cite{wang-aaai2020-cmdn} & 2020  & \checkmark &  & 60.7 & 52.5 & 40.5 & 23.5  & \multicolumn{1}{c|}{4.5} & \multicolumn{1}{c|}{33.3} & 62.3 & 53.1 & 74.2 & 58.7 & 31.6 & 4.7 & \multicolumn{1}{c|}{0.0} & \multicolumn{1}{c|}{30.1} & 55.4  & 57.6 \\
			Hui et al.\cite{hui-cvpr2021-cstm}& 2021 &  & \checkmark & 65.4 & 58.9 & 49.7 & 33.3 & \multicolumn{1}{c|}{9.1} & \multicolumn{1}{c|}{39.9} & 66.2 & 56.1 & 78.3 & 63.9 & 37.8 & 7.6 & \multicolumn{1}{c|}{0.0} & \multicolumn{1}{c|}{33.5} & 59.8 & 60.4 \\
			Ye et al.\cite{ye-pami2021-cmsa} & 2021  &  & \checkmark & 48.7 & 43.1 & 35.8 & 23.1  & \multicolumn{1}{c|}{5.2} & \multicolumn{1}{c|}{-} & 61.8 & 43.2 & 76.4 & 62.5 & 38.9 & 9.0 & \multicolumn{1}{c|}{0.0} & \multicolumn{1}{c|}{-} & 62.8 & 58.1 \\
			CMPC-V\cite{liu-pami2021-cmpc}& 2022  &  & \checkmark & 65.5 & 59.2 & 50.6 & 34.2 & \multicolumn{1}{c|}{9.8} & \multicolumn{1}{c|}{40.4} & 65.3 & 57.3 & 81.3 & 65.7 & 37.1 & 7.0 & \multicolumn{1}{c|}{0.0} & \multicolumn{1}{c|}{34.2} & 61.6 & 61.7 \\
			MTTR\cite{botach-cvpr2022-mttr} & 2022 & \checkmark & \checkmark & \underline{72.1} & \underline{68.4} & \underline{60.7} & \underline{45.6}  & \multicolumn{1}{c|}{\underline{16.4}} & \multicolumn{1}{c|}{\underline{44.7}} & \underline{70.2} & \underline{61.8} & \underline{91.0} & \underline{81.5} & \underline{57.0} & \underline{14.4} & \multicolumn{1}{c|}{\textbf{0.1}} & \multicolumn{1}{c|}{\underline{36.6}} & \underline{67.4} & \underline{67.9} \\
			\midrule[0.5pt]
			FTEA (Ours) & & \checkmark & \checkmark & \textbf{74.2} & \textbf{70.2} & \textbf{62.4} & \textbf{47.8}   & \multicolumn{1}{c|}{\textbf{17.1}} & \multicolumn{1}{c|}{\textbf{45.1}} & \textbf{71.2} & \textbf{62.8} & \multicolumn{1}{c}{\textbf{94.2}} & \multicolumn{1}{c}{\textbf{83.8}} & \multicolumn{1}{c}{\textbf{59.2}} & \multicolumn{1}{c}{\textbf{15.5}} & \multicolumn{1}{c|}{\textbf{0.1}} & \multicolumn{1}{c|}{\textbf{38.7}} & \multicolumn{1}{c}{\textbf{70.1}} & \multicolumn{1}{c}{\textbf{69.5}} \\ \toprule[0.75pt]
		\end{tabular}
	}
\end{table*}

\subsection{Comparison with State-of-the-art Methods}
We compare our FTEA approach with several state-of-the-art RVOS methods on benchmarks. The results on A2D Sentences and J-HMDB Sentences are recorded in Table~\ref{table:a2d_jhmdb} and the results on Ref-YouTube-VOS are shown in Table~\ref{table:ref-youtube-vos}. The best record is in boldface and the second best is underlined. Note that the model trained on A2D Sentences is directly used for segmenting all samples in J-HMDB Sentences. The results of those alternative methods are derived from their original papers, and we use the results of MTTR \cite{botach-cvpr2022-mttr} with the same temporal length 8 of one video clip. Note that we do not compare with the expensive model ReferFormer \cite{wu-cvpr2022-referformer} since it requires 8 NVIDIA V100 32GB GPUs, which are unavailable for most researchers. Generally, our FTEA method achieves consistently better segmentation performances than several strong baselines. Some discussions on the observations of comparison records are given below.

\textbf{A2D Sentences}. From the left part of Table~\ref{table:a2d_jhmdb}, FTEA outperforms SOTA alternatives across all evaluation metrics. In particular, FTEA has obvious gains of 2.1\% and 2.0\% over the best candidate MTTR on Precision@0.5 and Precision@0.8, respectively. Meanwhile, our method obtains the performance improvement of 1\% over MTTR in terms of both Overall IoU and Mean IoU. Both MTTR and FTEA adopt dynamic convolution and attention mechanism, and we attribute the performance gains to the fact that FTEA employs the newly developed stacked transformer and imposes the diversity constraint on the objective function, which can help the model to fuse multi-level semantics used for generating find the correct object from those candidate ones while providing more accurate pixel-level segmentation. Besides, the methods without dynamic convolution such as CMPC-V \cite{liu-pami2021-cmpc} performs much worse than MTTR and FTEA, which demonstrates the effectiveness of generating a group of convolution kernels conditioned on different text queries and video frames.  

\textbf{J-HMDB Sentences}. From the right part of Table~\ref{table:a2d_jhmdb}, FTEA derives largely better segmentation results in comparison with other methods. For example, it achieves 3.2\% gain on Precision@0.5, 2.1\% gain on mAP, and 2.7\% gain in terms of Overall IoU, compared to the strongest baseline MTTR \cite{botach-cvpr2022-mttr}. In other words, more obvious performance boosting has been observed on J-HMDB Sentences than that on A2D Sentences. The former has 21 action classes while the latter has only 8 ones, which suggests that our method can better handle videos with more actions. This also consolidates that the fully transformer-equipped architecture exhibits clear advantages for referring video object segmentation task. Note that all methods get very poor performance on Precision@0.9, which might be because the model is trained on A2D Sentences and the ground-truth masks in J-HMDB Sentences are generated by a coarse human puppet model, leading to some inaccuracy.

\begin{table}[!t]
	\centering
	\caption{Comparisons on Ref-YouTube-VOS validation set.}
	\label{table:ref-youtube-vos}
	\small
	\setlength{\tabcolsep}{0.4mm}{
		\begin{tabular}{lccccccccc}
			\toprule[0.75pt]
			Method & Year &  DC & Att & Visual Enc. & Text Enc. & $\mathcal{J}$ & $\mathcal{F}$ & $\mathcal{J}$\&$\mathcal{F}$ \\ \midrule[0.5pt]
			URVOS\cite{seo-eccv2020-urvos} & 2020 &  & \checkmark & ResNet50 & Word emb. & 45.3 & 49.2 & 47.2 \\
			CMPC-V\cite{liu-pami2021-cmpc} & 2022  & & \checkmark & I3D & GloVe & 45.6 & 49.3 & 47.5 \\
			SynthRef\cite{kazakos-naaclw2021-synthref} & 2021  &  & & DeepLabv3 & BERT & 39.5 & - & - \\
			Ding et al.\cite{ding-arxiv2021-pmin} & 2021  & & \checkmark & ResNeSt & GRU & \underline{53.7} & \underline{56.0} & \underline{54.8} \\
			MTTR\cite{botach-cvpr2022-mttr} & 2022  & \checkmark & \checkmark & Swin-T & RoBERTa & - & - & 54.6 \\
			MTTR*\cite{botach-cvpr2022-mttr} & 2022 & \checkmark & \checkmark & Swin-T & RoBERTa & 52.1 & 55.2 & 53.6 \\
			\midrule[0.5pt]
			FTEA (Ours) &  & \checkmark & \checkmark & Swin-T & RoBERTa & \textbf{55.0} & \textbf{58.0} & \textbf{56.5} \\
			\toprule[0.75pt]
		\end{tabular}
	}
\end{table}

\textbf{Ref-YouTube-VOS}. As can be seen from Table~\ref{table:ref-youtube-vos}, our approach gets the most promising performance among several SOTA methods including some ensemble approach like \cite{ding-arxiv2021-pmin} trained on additional data sets. For instance, the ensemble method \cite{ding-arxiv2021-pmin} obtains lower performance by 2\% on Contour Accuracy ($\mathcal{F}$) compared to ours, while MTTR lowers the performance by 1.9\% in terms of $\mathcal{J} \& \mathcal{F}$. Note that MTTR with a star in the superscript indicates that the results are reproduced using the released GitHub code when the temporal length of clip equals 8. FTEA not only surpasses the strong baseline MTTR, but also significantly outperforms the ensemble method (non end-to-end) which is the second best. This demonstrates that the stacked transformer design and the imposed diversity loss explicitly contribute to positive performance gains, while there are thousands of objects and text expressions.

\subsection{Ablation Studies}
We conduct ablation studies on the benchmark test sets to examine the influences of each component in FTEA, including stacked transformer, and diversity loss. Unless stated otherwise, we only change the component to be examined while keeping others fixed. 

\begin{table}[!t]
	\centering
	\caption{Ablation study of FTEA components on A2D Sentences test set. ST: Stacked Transformer; DL: Diversity Loss. }
	\label{table:abl-blocks-a2d}
	\small
	\setlength{\tabcolsep}{0.7mm}{
		\begin{tabular}{ccc|ccccc|c|cc}
			\toprule[0.75pt]
			\multirow{2}{*}{Baseline} & \multirow{2}{*}{ST} & \multirow{2}{*}{DL} & \multicolumn{5}{c|}{Precision} & mAP & \multicolumn{2}{c}{IoU} \\ \cline{4-11} 
			&  &  & 0.5 & 0.6 & 0.7 & 0.8 & 0.9 & 0.5:0.95 & Overall  & Mean  \\ \midrule[0.5pt]
			\checkmark &  &  & 72.1 & 68.4 & 60.7 & 45.6 & 16.4 & 44.7 & 70.2 & 61.8 \\
			\checkmark & \checkmark  & & \underline{73.9} & \textbf{70.2} & 61.8 & \underline{46.9} & 16.1 & \underline{44.9} & 70.6 & \underline{62.7} \\
			\checkmark &  &\checkmark  & 73.2 & 69.2 & \underline{62.1} & 46.8 & \underline{16.9} & 44.8 & \underline{70.7} & 62.5 \\
			\checkmark & \checkmark & \checkmark & \textbf{74.2} & \textbf{70.2} & \textbf{62.4} & \textbf{47.8} & \textbf{17.1} & \textbf{45.1} & \textbf{71.2} & \textbf{62.8} \\
			\toprule[0.75pt]
		\end{tabular}
	}
\end{table}

\begin{table}[!t]
	\centering
	\caption{Ablation study of components on J-HMDB Sentences. }
	\label{table:abl-blocks-jhmdb}
	\small
	\setlength{\tabcolsep}{0.7mm}{
		\begin{tabular}{ccc|ccccc|c|cc}
			\toprule[0.75pt]
			\multirow{2}{*}{Baseline} & \multirow{2}{*}{ST} & \multirow{2}{*}{DL} & \multicolumn{5}{c|}{Precision} & mAP & \multicolumn{2}{c}{IoU} \\ \cline{4-11} 
			&  &  & 0.5 & 0.6 & 0.7 & 0.8 & 0.9 & 0.5:0.95 & Overall  & Mean  \\ \midrule[0.5pt]
			\checkmark &  &  & 91.0 & 81.5 & 57.0 & 14.4 & 0.1 & 36.6 & 67.4 & 67.9 \\
			\checkmark & \checkmark & &\underline{93.6} & \underline{83.4} & \underline{58.8} & \underline{15.2} & 0.1 & \underline{38.0} & \underline{69.6} & \underline{69.5} \\
			\checkmark &  & \checkmark & 93.3 & 83.0  & 58.2 & 15.0 & 0.1 & 37.7 & 68.8 & 68.9 \\	
			\checkmark & \checkmark & \checkmark & \textbf{94.2} & \textbf{83.8} & \textbf{59.2} & \textbf{15.5} & 0.1 & \textbf{38.7} & \textbf{70.1} & \textbf{69.5} \\
			\toprule[0.75pt]
		\end{tabular}
	}
\end{table}

\begin{table}[!t]
	\centering
	\caption{Ablation study of components on Ref-YouTube-RVOS. }
	\label{table:abl-blocks-ytbvos}
	\small
	\setlength{\tabcolsep}{1.5mm}{
		\begin{tabular}{ccc|lll}
			\toprule[0.75pt]
			Baseline & ST & DL & $\mathcal{J}$ & $\mathcal{F}$ &  $\mathcal{J}$\&$\mathcal{F}$  \\ \midrule[0.5pt]
			\checkmark &            &            & 52.1          & 55.2          & 53.6    \\
			\checkmark & \checkmark &            & 53.6 (+1.5)   & 56.8 (+1.6)   & 55.2 (+1.6) \\
			\checkmark &            & \checkmark & 53.3 (+1.2)   & 56.5 (+1.3)   & 54.9 (+1.3) \\
			\checkmark & \checkmark & \checkmark & \textbf{55.0} (+2.9) & \textbf{58.0} (+2.8) & \textbf{56.5} (+2.9) \\
			\toprule[0.75pt]
		\end{tabular}
	}
\end{table}

\textbf{Components of FTEA}. We first evaluate the effectiveness of stacked transformer and diversity loss, and the results are shown in Table~\ref{table:abl-blocks-a2d}, \ref{table:abl-blocks-jhmdb}, and \ref{table:abl-blocks-ytbvos}. Here, the baseline model is MTTR \cite{botach-cvpr2022-mttr} which uses convolution based Feature Pyramid Network (FPN) \cite{lin-cvpr2017-fpn} as mask decoder and trained without diversity loss. DL and ST denote Diversity Loss and Stacked Transformer, respectively.  On the one hand, when using our designed stacked transformer as mask decoder, the Precision@0.5 is upgraded from 72.1\% to 73.9\% on A2D Sentences, and from 91.0\% to 93.6\% on J-HMDB Sentences. Simultaneously, Mean IoU is boosted by 0.9\% on A2D Sentences and 1.6\% on J-HMDB Sentences. This explicitly shows the merits of stacked architecture. On the other hand, when adding the diversity loss during the baseline model training on A2D Sentences and J-HMDB Sentences, it has a performance gain of 1.4\% and 1.8\% on Precision@0.7, respectively. This verifies that the diversity loss enables encouraging the diversity among candidate object masks so as to alleviate the problem of appearance similarity of candidate objects. While unifying the two components together in fully transformer-equipped framework, it gets accumulative performance gains on A2D Sentences and J-HMDB Sentences, \eg, 2.1\% and 3.2\% on Precision@0.5, respectively. In addition, even larger performance gaps are observed on Ref-YouTube-RVOS as in Table~\ref{table:abl-blocks-ytbvos}. This shows the additive positive influences of the two components on the referring segmentation performance.

\textbf{Stacked Transformer}. We examine different modules available for substituting Stacked Transformer as mask decoder, and the results are shown in Table~\ref{table:abl-stackTrans-a2d} and Table~\ref{table:abl-stackTrans-jhmdb}. We keep all the other settings fixed while only changing the building block of Mask Decoder (MD). Compared with the commonly used FPN \cite{lin-cvpr2017-fpn}, adding the Stacked Attention (SA) enhances the performance of Feed Forward Network (FFN), which is for the reason that it exploits the object-level spatial context by applying candidate object kernels on the pixel-wise feature maps. While further adopting Stacked FFN, the segmentation performance gets additional improvements. This is because Stacked FFN incorporates the group-wise multi-layer perceptron to learn the appearance attributes of each candidate object independently. Moreover, due to the group sparsity by adopting group linear mapping, the number of model parameters in Stacked Transformer is only one-fifth of that of previously used Feature Pyramid Network. In addition, we explore the three components SA, FFN, SFFN individually for the in-depth analysis in Table~\ref{table:abl-stackTrans-a2d}. As shown in the bottom group, SA brings the most significant gain and achieves 44.8\% in terms of mAP, but simply only using SFFN or FFN degenerates the performance a lot. Meanwhile, SFFN performs a bit better than FFN with the much less parameters. This further validate the advantages of cooperatively unifying Stacked Attention and Stacked Feed Forward Network. Hence, we can draw a conclusion that SA plays the most important role in contributing to the success of Stacked Transformer module. 

\begin{table}[!t]
	\centering
	\caption{Ablation study of Stacked Transformer on A2D-Sentences test set. FPN: Feature Pyramid Network; SA: Stacked Attention; SFFN: Stacked Feed Forward Network.}
	\label{table:abl-stackTrans-a2d}
	\small
	\setlength{\tabcolsep}{0.7mm}{
		\begin{tabular}{lc|ccccc|c|cc}
			\toprule[0.75pt]
			\multirow{2}{*}{ST} & \multicolumn{1}{l|}{\multirow{2}{*}{\#Para.}} & \multicolumn{5}{c|}{Precision} & mAP & \multicolumn{2}{c}{IoU} \\ \cline{3-10} 
			& \multicolumn{1}{l|}{} & 0.5 & 0.6 & 0.7 & 0.8 & 0.9 & 0.5:0.95 & Overall & Mean \\ \midrule[0.5pt]
			FPN & 1.0M & 73.2 & \underline{69.2} & 62.1 & 46.8 & 16.9 & 44.7 & 70.7 & 62.5 \\
			SA+FFN & 0.5M & \underline{73.6} & \underline{69.2} & \underline{62.2} & \underline{47.1} & \underline{17.0} & \underline{44.9} & \underline{70.8} & \underline{62.6} \\
			SA+SFFN & \textbf{0.2M} & \textbf{74.2} & \textbf{70.2} & \textbf{62.4} &\textbf{47.8} & \textbf{17.1} & \textbf{45.1} & \textbf{71.2} & \textbf{62.8} \\
			\midrule[0.5pt]
			SA             & 0.1M     & 73.5    &  69.1    & 61.4  & 44.9      & 15.5  & 44.8   & 70.7    & 62.0\\
			SFFN           & 0.05M     & 71.7    &  66.4    & 56.5  & 40.1      & 12.3  & 41.0   & 68.4    & 59.9\\
			FFN            & 0.4M    & 70.8    &  65.6    & 56.1  & 39.4      & 12.7  & 40.0   & 67.7    & 59.2\\
			\toprule[0.75pt]
		\end{tabular}
	}
\end{table}

\begin{table}[!t]
	\centering
	\caption{Ablation study of Stacked Transformer on J-HMDB Sentences.}
	\label{table:abl-stackTrans-jhmdb}
	\small
	\setlength{\tabcolsep}{0.7mm}{
		\begin{tabular}{lc|ccccc|c|cc}
			\toprule[0.75pt]
			\multirow{2}{*}{MD} & \multicolumn{1}{l|}{\multirow{2}{*}{\#Para.}} & \multicolumn{5}{c|}{Precision} & mAP & \multicolumn{2}{c}{IoU} \\ \cline{3-10} 
			& \multicolumn{1}{l|}{} & 0.5 & 0.6 & 0.7 & 0.8 & 0.9 & 0.5:0.95 & Overall & Mean \\ \midrule[0.5pt]
			FPN & 1.0M & 91.0 & 81.5 & 57.0 & 14.4 & 0.1 & 36.6 & 67.4 & 67.9 \\
			SA+FFN & 0.5M & \underline{92.2} 
			& \underline{82.8} & \underline{58.4} & \underline{15.0} & 0.1 & \underline{37.9} & \underline{68.8} & \underline{68.7} \\
			SA+SFFN & $\mathbf{0.2M}$ & $\mathbf{94.2}$ & \textbf{83.8} & \textbf{59.2} & \textbf{15.5} & 0.1 & \textbf{38.7} & \textbf{70.1} & \textbf{69.5} \\
			\toprule[0.75pt]
		\end{tabular}
	}
\end{table}

\begin{table}[!t]
	\centering
	\caption{Ablation study of diversity loss variants on A2D Sentences. }
	\label{table:abl-div-loss-a2d}
	\small
	\setlength{\tabcolsep}{1.1mm}{
		\begin{tabular}{l|ccccc|c|cc}
			\toprule[0.75pt]
			\multirow{2}{*}{$\mathcal{L}_{\text{div}}$} & \multicolumn{5}{c|}{Precision} & mAP & \multicolumn{2}{c}{IoU} \\ \cline{2-9}
			& 0.5 & 0.6 & 0.7 & 0.8 & 0.9 & 0.5:0.95 & Overall & Mean \\ \midrule[0.5pt]
			Base & \underline{73.6} & \underline{69.8} & 62.3 & 47.1 & 16.8 & 44.7 & \underline{70.9} & 62.4 \\
			+ $\ell_1$  & \textbf{74.2} & \textbf{70.2} & \textbf{62.4} &\textbf{47.8} & \textbf{17.1} & \textbf{45.1} & \textbf{71.2} & \textbf{62.8} \\
			+ $\ell_2$  & 73.2 & 69.2 & \textbf{62.4} & \textbf{47.8} & \underline{16.9} & \underline{44.7} & 70.6 & \underline{62.5} \\
			+ Fro  & 72.6 & 69.1 & 62.2 & 46.7 & 16.5 & 44.6 & 70.2 & 62.0 \\
			\toprule[0.75pt]
		\end{tabular}
	}
\end{table}

\begin{table}[!t]
	\centering
	\caption{Ablation study of diversity loss variants on J-HMDB Sentences. }
	\label{table:abl-div-loss-jhmdb}
	\small
	\setlength{\tabcolsep}{1.1mm}{
		\begin{tabular}{l|ccccc|c|cc}
			\toprule[0.75pt]
			\multirow{2}{*}{$\mathcal{L}_{\text{div}}$} & \multicolumn{5}{c|}{Precision} & mAP & \multicolumn{2}{c}{IoU} \\ \cline{2-9}
			& 0.5 & 0.6 & 0.7 & 0.8 & 0.9 & 0.5:0.95 & Overall & Mean \\ \midrule[0.5pt]
			Base & \underline{93.6} & \underline{83.2} & \underline{58.9} & \underline{15.2} & 0.1 & \underline{38.3} & \underline{69.6} & \underline{69.5}\\
			+ $\ell_1$  & \textbf{94.2} & \textbf{83.8} & \textbf{59.2} &\textbf{15.5} & 0.1 & \textbf{38.7} & \textbf{70.1} & \textbf{69.6} \\
			+ $\ell_2$  & 93.4 & 83.0 & \underline{58.9} & 15.1 & 0.1 & 38.2 & 69.3 & 69.3 \\
			+ Fro  & 93.0 & 82.7 & 58.5 & 14.9 & 0.1 & 38.0 & 68.7 & 68.9 \\
			\toprule[0.75pt]
		\end{tabular}
	}
\end{table}
%

\begin{table}[!t]
	\centering
	\caption{Ablation study of hyper-parameter $\lambda_{\text{div}}$ on A2D Sentences. }
	\label{table:abl-lamda}
	\small
	\setlength{\tabcolsep}{1.5mm}{
		\begin{tabular}{l|ccccc|c|cc}
			\toprule[0.75pt]
			\multirow{2}{*}{$\lambda_{\text{div}}$} & \multicolumn{5}{c|}{Precision} & mAP & \multicolumn{2}{c}{IoU} \\ \cline{2-9} 
			& 0.5 & 0.6 & 0.7 & 0.8 & 0.9 & 0.5:0.95 & Overall & Mean \\ \midrule[0.5pt]
			0.01 & 71.7 & 67.8 & 60.8 & 46.1 & \underline{16.8} & 43.7 & 70.0 & 61.6 \\
			0.03 & 72.1 & 68.3 & 61.1 & 46.3 & 16.4 & 44.0 & \underline{70.3} & 61.6 \\
			0.05 & \underline{73.6} & \underline{69.1} & \underline{61.7} & \underline{46.8} & 16.1 & \underline{44.6} & 70.2 & \underline{62.0} \\
			0.07 & \textbf{74.2} & \textbf{70.2} & \textbf{62.4} &\textbf{47.8} & \textbf{17.1} & \textbf{45.1} & \textbf{71.2} & \textbf{62.8} \\
			0.09 & 72.7 & 68.7 & 60.3 & 46.3 & 16.4 & 44.2 & 69.5 & 61.3 \\
			\toprule[0.75pt]
		\end{tabular}
	}
\end{table}
%

%
\begin{table}[!t]
	\centering
	\caption{Ablation study of candidate object number on A2D Sentences. }
	\label{table:abl-objectnum}
	\small
	\setlength{\tabcolsep}{1.5mm}{
		\begin{tabular}{l|ccccc|c|cc}
			\toprule[0.75pt]
			\multirow{2}{*}{$K$} & \multicolumn{5}{c|}{Precision} & mAP & \multicolumn{2}{c}{IoU} \\ \cline{2-9} 
			& 0.5 & 0.6 & 0.7 & 0.8 & 0.9 & 0.5:0.95 & Overall & Mean \\ \midrule[0.5pt]
			10 & 72.1 & 67.4 & 59.3 & 43.2 & 13.5 & 40.3 & 69.4 & 61.0 \\
			30 & 72.7 & 67.7 & 60.3 & 45.0 & 14.5 & 41.6 & 69.9 & 61.7 \\
			50 & \textbf{74.2} & \textbf{70.2} & \textbf{62.4} &\textbf{47.8} & \textbf{17.1} & \textbf{45.1} & \textbf{71.2} & \textbf{62.8} \\
			70 & \underline{73.2} & \underline{69.2} & \underline{61.9} & \underline{47.2} & \underline{16.5} & \underline{44.8} & \underline{70.2} & \underline{62.5} \\
			90 & 73.1 & 68.8 & 60.9 & 46.0 & 15.3 & 41.8 & 69.8 & 62.1 \\
			\toprule[0.75pt]
		\end{tabular}
	}
\end{table}

\begin{figure*}[!t]
	\centering
	\includegraphics[width=0.9\linewidth]{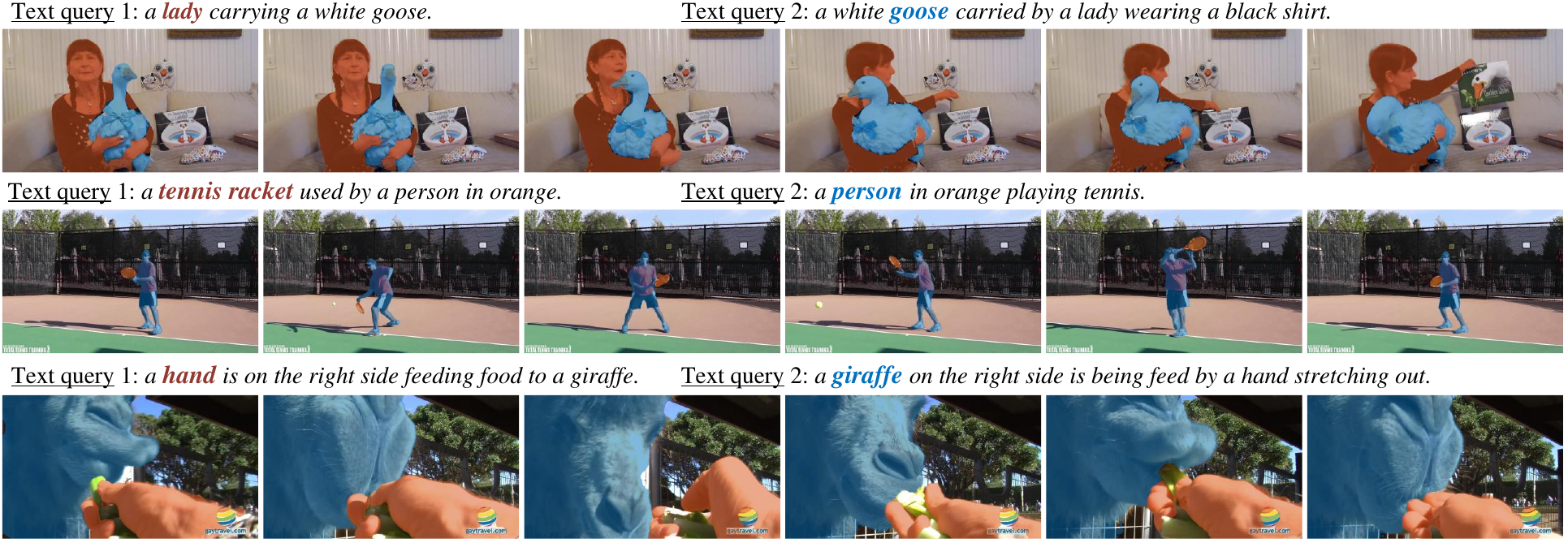}
	\caption{Qualitative results on Ref-YouTube-VOS under adverse scenarios. Mask and text color indicates some object in video clip.}
	\label{fig:visualize-youtube}
\end{figure*}

\begin{figure}[!t]
	\centering
	\includegraphics[width=0.45\textwidth]{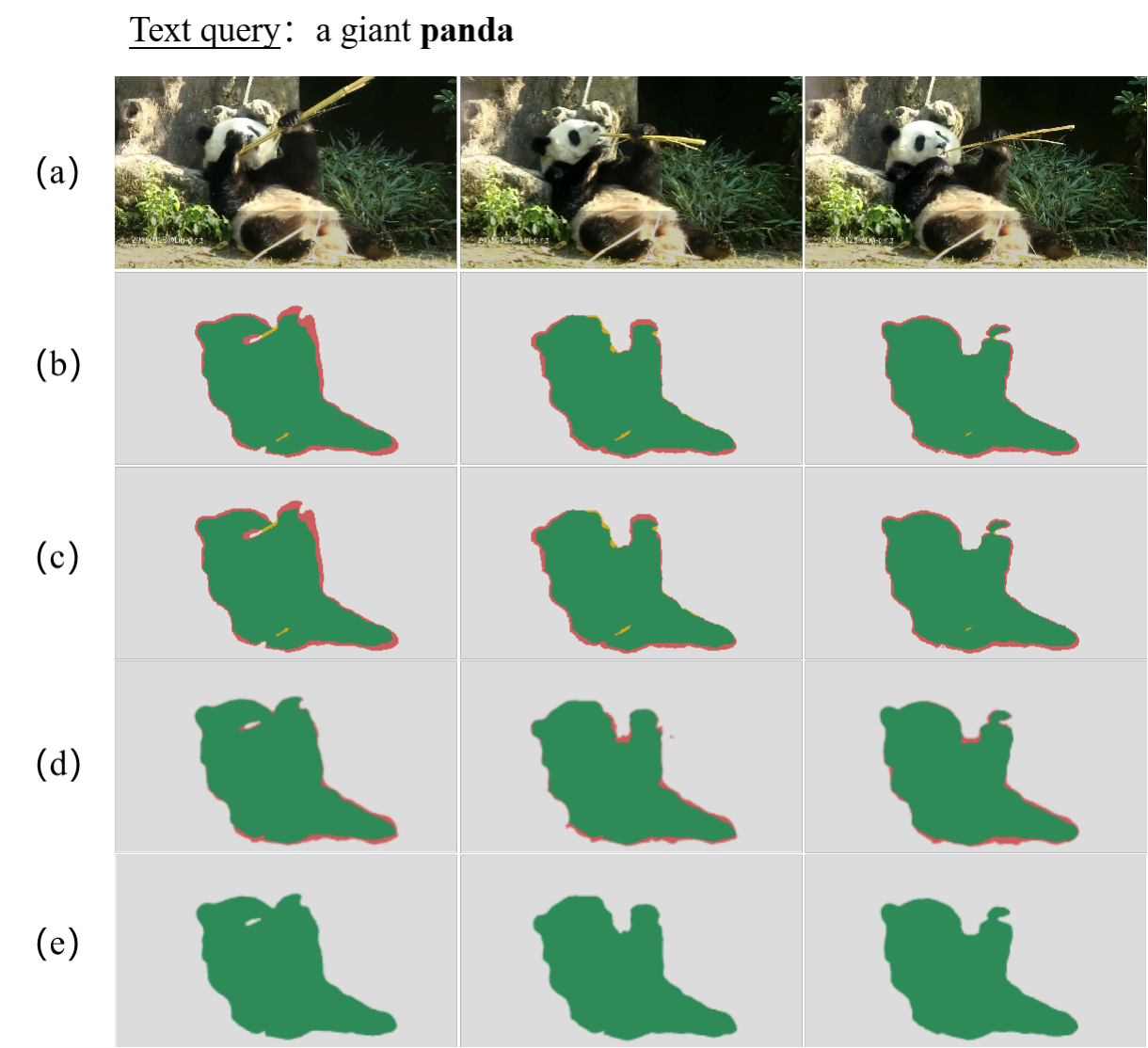}
	\includegraphics[width=0.45\textwidth]{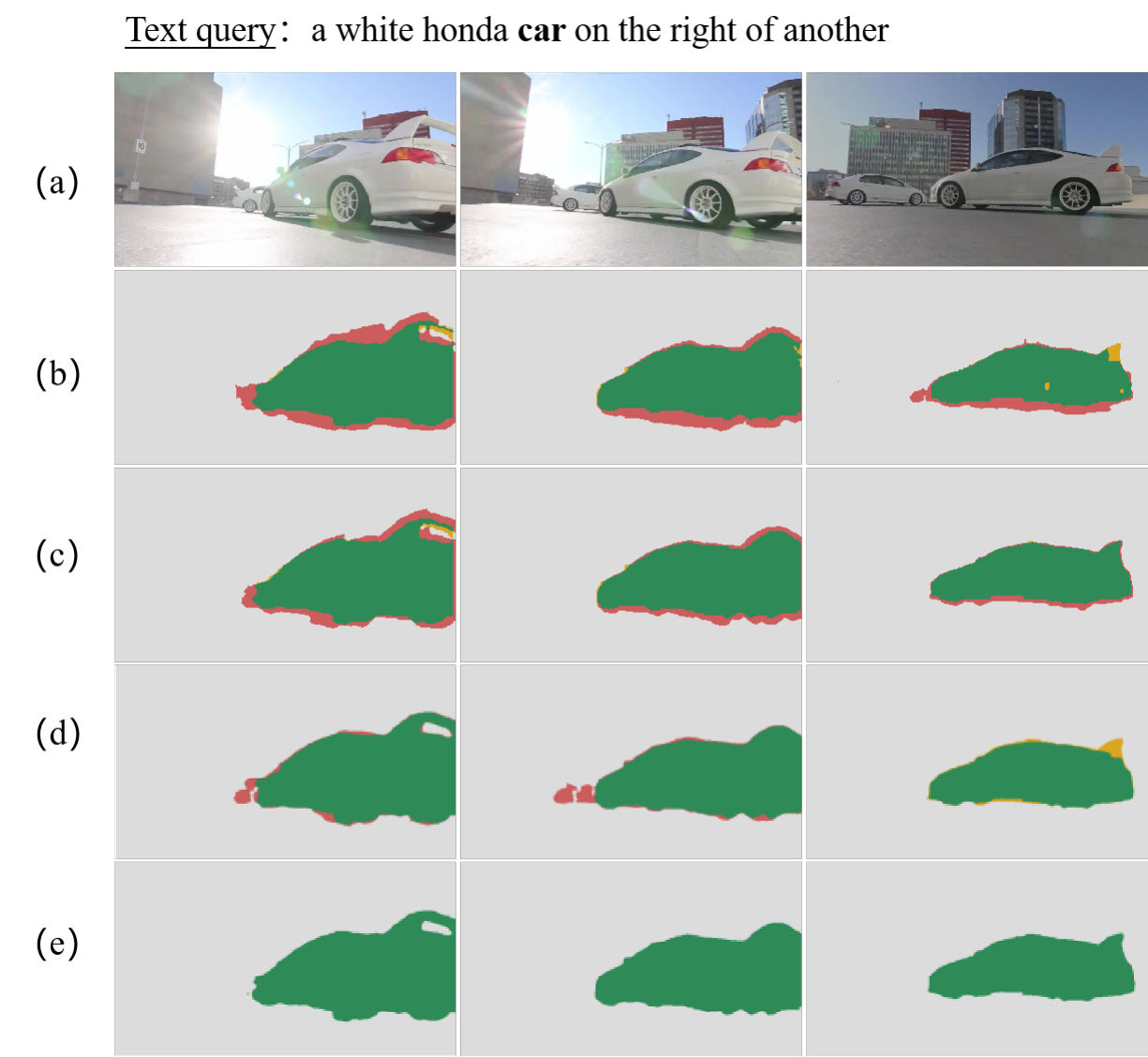}		
	\caption{Qualitative comparison between the baselines and our model on Ref-YouTube-VOS validation set. (a) Video frames; (b) URVOS \cite{seo-eccv2020-urvos}; (c) CMPC-V \cite{liu-pami2021-cmpc}; (d) MTTR \cite{botach-cvpr2022-mttr}; (e) FTEA (Ours).
	\label{fig:viscompare-youtube}}
\end{figure}

\textbf{Diversity loss variant}. To explore the effect of the norm style imposed on the diversity term, we show the results of using $\ell_1$-norm, $\ell_2$-norm, and Frobenius norm, in Table~\ref{table:abl-div-loss-a2d} and \ref{table:abl-div-loss-jhmdb}. The first row (Base) show the result of diversity loss without any normalization term. The different norms are applied on the learnable matrix $\mathbf{W}_{\text{div}}$, which is used for projecting the candidate object kernel into subspace where the object-wise similarity among all candidate objects are computed. From the table, the performance of using $\ell_1$-norm achieves the best. The possible reason is that there are some redundant candidate objects in produced mask sequences and $\ell_1$-norm introduces sparsity constraint onto the diversity loss for reducing redundancy, so as to alleviate the overfitting problem during model optimization.

\textbf{Diversity loss hyper-parameter}. To examine the contribution of the diversity loss term to the model on A2D Sentences test set, we vary the hyper-parameter $\lambda_{\text{div}}$ from 0.01 to 0.09 with an interval of 0.02 and the results are shown in Table~\ref{table:abl-lamda}. As can be seen from the table, the performance goes up in the beginning, peaks at 0.07, and then goes down. So we set $\lambda_{\text{div}}$ to 0.7 through all experiments. Unlike Dice loss \cite{milletari-3dv2016-dice} and Focal loss \cite{lin-iccv2017-focal}, which calculate the loss of candidate object individually, our diversity loss is applied on candidate objects to calculate the similarity of object pairs, resulting in greater loss values. So the balancing coefficient $\lambda_{\text{div}}$ is small.

\textbf{Candidate object number}. To examine the influences of candidate object number, we vary the number $K$ from 10 to 90 with a gap of 20 and show the segmentation results on A2D Sentences test set in Table~\ref{table:abl-objectnum}. As observed in the table, the performance rises by 4.8\% when $K$ changes from 10 to 50 and drops by 3.3\% from 50 to 90, in terms of mAP. Neither too many candidate objects nor too less ones can achieve good performance. Hence, we choose 50 for $K$ since the performance saturates at this point. This phenomenon stems from the fact that the actual number of objects in video is much smaller than that of candidate objects, and the model will be trapped in local minima when the candidate object number gets very large. Moreover, too small number does not allow the model to capture the diverse pattern of different candidate objects in video.

\subsection{Qualitative Analysis}
To quantitatively demonstrate the robustness of our FTEA method, we illustrate the segmentation results of three randomly selected video clips from Ref-YouTube-VOS validation set under adverse scenarios in Figure~\ref{fig:visualize-youtube}. From this figure, we can see that our model can successfully locate the text referred objects while producing accurate masks in several challenging video scenes, such as object occlusions(\ie, lady hands before goose at top row), severe appearance changes (\ie, sportsman body pose variation at middle row), partial appearance missing (\ie, large part of giraffe is unseen at bottom row). This might because our method can well capture object-level spatial context by using fully transformer-equipped framework with additionally imposed diversity loss term.

Moreover, to give an intuitive visualization comparison, we randomly selected two video clips from Ref-YouTube-VOS validation set and show the segmentation results in Figure~\ref{fig:viscompare-youtube}. For both left and right clips in this figure, the first row shows the sampled frames in video clip, the three middle rows show the results of baseline models URVOS \cite{seo-eccv2020-urvos}, CMPC-V \cite{liu-pami2021-cmpc}, and MTTR \cite{botach-cvpr2022-mttr}, respectively; the bottom row is the result of our FTEA model. Since the ground-truth mask is unavailable, we take the results of FTEA as the reference and highlight the false positive pixels in red and false negative pixels in yellow for the baseline models. As vividly shown, our method produces the smoother contours of the giant panda and the while car while the baseline models may treat some surrounding background pixels as object part by mistake. For example, ours can well handle car appearance variation and the baselines do not perform satisfactorily. This can be due to the fact that our method adopts the strategy of progressive object learning, which finds the target object from a group of candidate objects by stacked transformers.

\begin{figure}[!h]	
	\centering	
	\includegraphics[width=0.5\textwidth]{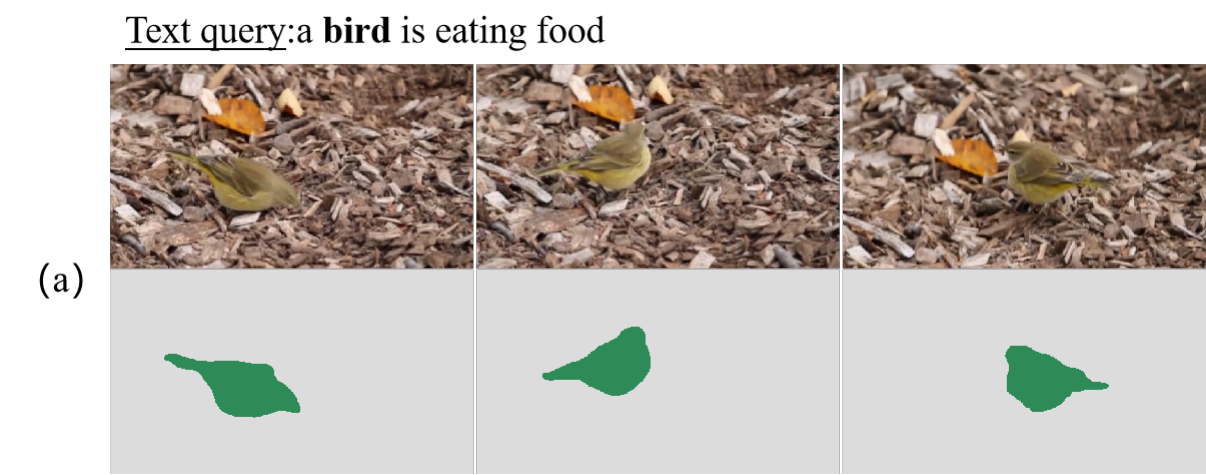}	
	\includegraphics[width=0.5\textwidth]{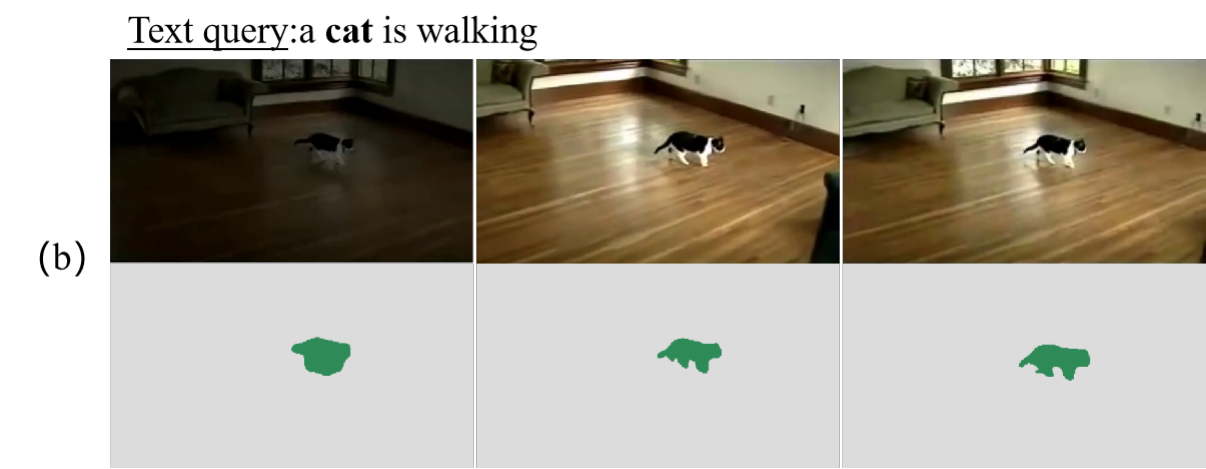}
	\includegraphics[width=0.5\textwidth]{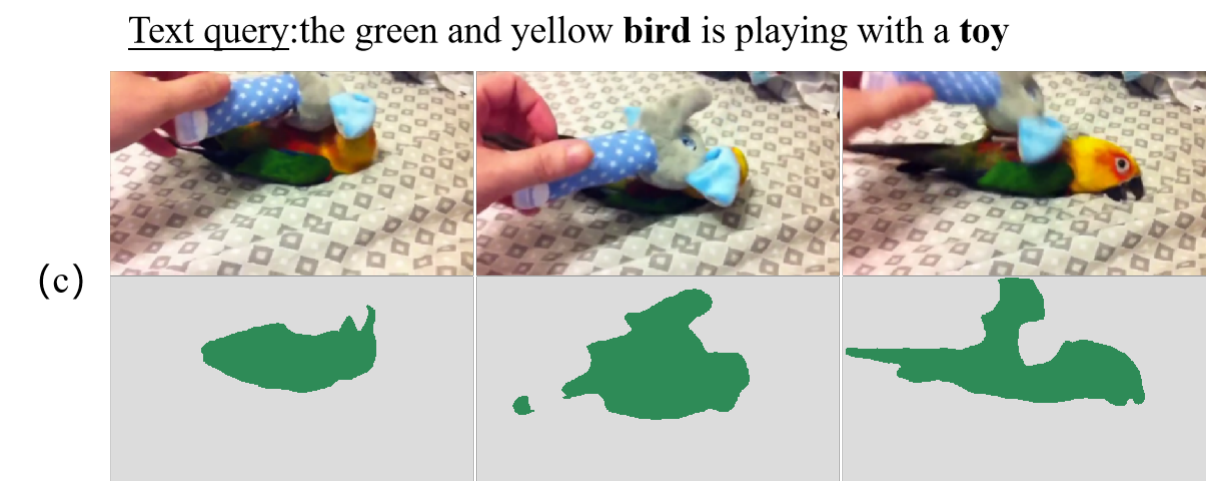}			
	\caption{Failure cases of our model on A2D Sentences test set.}
	\label{fig:failurecase}
\end{figure}

In addition, we show several failure cases by randomly selecting several video clips from A2D Sentences test set. As show in Fig.~\ref{fig:failurecase}, the bird mouth is hard to segment in the top clip when the mouth is very small and the body shares the similar color with the background; the cat mask in the beginning of the middle clip is obscure when the lighting condition is very poor; the masks of the bird and the toy are confused in the bottom clip when the two objects are overlapped. This will inspire our future research works of addressing these issues.

\section{Discussion}
\label{discuss}
Unlike traditional VOS tasks, RVOS requires to segment the objects referred by natural language queries, which desires to bridge the semantic gap between video and language. Essentially, we should model the pixel-wise spatiotemporal relations among the frames in video, and align the target area with the referred object. Existing methods usually regard RVOS as a pixel-wise classification problem which classifies the pixels into foreground (objects) or background (scenes), failing to well consider the object spatial relations. Instead, we treat RVOS as an object-wise classification problem, \ie, capturing the appearances of the objects referred at the same time, thus modeling their spatial relations. 

For the candidate object location, existing methods often adopt the convolution and upsampling to decode the features into masks. While these frame features capture the local context, they fail to model the global relations. This inspires us to develop a transformer-based network for decoding the features into object masks, by capturing the globally spatial context. Moreover, our framework completely adopts the stacked transformers to encode both the visual and the text features, which has been demonstrated by the above experiments. 

In addition, there is a fact that the number of candidate object masks is much more than that of objects in video, and one object will be aligned with multiple candidate masks, which remains under-explored by previous works. So, we propose to impose the diversity loss on candidate objects to make them diverse as much as possible, so that the objects in video can be maximally identified. The effectiveness of this idea has been validated by the ablation records in Table~\ref{table:abl-div-loss-a2d} and \ref{table:abl-div-loss-jhmdb}.

In fact, our transformation block, \ie, Stacked Transformer, can be applied to other pixel-wise segmentation tasks, while the diversity loss can be applied in DETR-like architecture for other tasks, such as object detection. From the practical view, our RVOS model can be used in video editing, automobile driving, and human-robot interaction. For example, it facilitates the production of the customized film by segmenting the designated object or person quickly from thousands of source videos.

\section{Conclusion}
\label{conclusion}
This paper has proposed a fully transformer-equipped architecture for referring video object segmentation, termed FTEA, which can be trained in an end-to-end manner. Unlike existing methods failing to well capture long-range spatial-temporal relations across video frames, our model completely use transformers for visual and text feature extraction, cross-modal alignment, and mask decoding. In particular, we developed the stacked transformers for composing mask decoder to better capture object-level spatial context, which is beneficial for identifying the target object from candidate objects. Furthermore, we impose the diversity regularization term on the objective function to promote yielding diverse object masks, such that as many as objects in video are taken into account by the model. To examine the performance of our method, we conducted a lot of experiments on three benchmark databases, and the extensive results have clearly validated the superiority of our FTEA model in end-to-end RVOS task. 

However, there are still some limitations of our method. First, it may fail to identify the complete contours of referred objects in some adverse scenarios, such as similar appearance of object and background, poor lighting condition, and object overlapping. Second, it still requires large computational sources for training and inference, which will prohibit its widespread applications. In future, we will explore the way of facilitating the inference of language relations among multiple objects in similar appearance, and attempt to compress the model by adopting the knowledge distillation or pruning strategy.

\small

\end{document}